\documentclass[11pt]{article}

\usepackage[]{acl}

\usepackage{times}
\usepackage{latexsym}
\usepackage{graphicx}
\usepackage{booktabs}
\usepackage{makecell}   
\usepackage{booktabs}
\usepackage{multirow}
\usepackage{amsmath}
\usepackage{amsfonts}

\usepackage[inline]{enumitem}

\usepackage[T1]{fontenc}

\usepackage[utf8]{inputenc}

\usepackage{microtype}

\usepackage{inconsolata}

\usepackage[most]{tcolorbox}  
\usepackage{lipsum}
\usepackage{fvextra}          

\usepackage[capitalize]{cleveref}
\crefname{appsection}{Appendix}{Appendices}

\newtcolorbox{fullpromptbox}[1][]{
  breakable,
  enhanced,
  colback=gray!10,
  colframe=black!40,
  fontupper=\ttfamily\scriptsize,
  width=\textwidth,
  boxrule=0.4pt,
  arc=3pt,
  left=5pt,
  right=5pt,
  top=5pt,
  bottom=5pt,
  sharp corners,
  attach boxed title to top left={yshift=-1mm, xshift=4mm},
  title=#1,
}

\title{LAD-RAG: Layout-aware Dynamic RAG for Visually-Rich Document Understanding}

\author{
  Zhivar Sourati\textsuperscript{\rm 1,2},
  Zheng Wang\textsuperscript{\rm 2},
  Marianne Menglin Liu\textsuperscript{\rm 2},
  Yazhe Hu\textsuperscript{\rm 2},
  Mengqing Guo\textsuperscript{\rm 2}
  \AND
  Sujeeth Bharadwaj\textsuperscript{\rm 2},
  Kyu Han\textsuperscript{\rm 2},
  Tao Sheng\textsuperscript{\rm 2},
  Sujith Ravi\textsuperscript{\rm 2},
  Morteza Dehghani\textsuperscript{\rm 1} \and
  Dan Roth\textsuperscript{\rm 2} \\
  \textsuperscript{\rm 1}University of Southern California \\
  \textsuperscript{\rm 2}Oracle AI
}

\begin{document}
\maketitle
\begin{abstract}
Question answering over visually rich documents (VRDs) requires reasoning not only over isolated content but also over documents' structural organization and cross-page dependencies. However, conventional retrieval-augmented generation (RAG) methods encode content in isolated chunks during ingestion, losing structural and cross-page dependencies, and retrieve a fixed number of pages at inference, regardless of the specific demands of the question or context. This often results in incomplete evidence retrieval and degraded answer quality for multi-page reasoning tasks. To address these limitations, we propose LAD-RAG, a novel Layout-Aware Dynamic RAG framework. During ingestion, LAD-RAG constructs a symbolic document graph that captures layout structure and cross-page dependencies, adding it alongside standard neural embeddings to yield a more holistic representation of the document. During inference, an LLM agent dynamically interacts with the neural and symbolic indices to adaptively retrieve the necessary evidence based on the query. Experiments on MMLongBench-Doc, LongDocURL, DUDE, and MP-DocVQA demonstrate that LAD-RAG improves retrieval, achieving over 90\% perfect recall on average without any top‑$k$ tuning, and outperforming baseline retrievers by up to 20\% in recall at comparable noise levels, yielding higher QA accuracy with minimal latency. 
\end{abstract}

\section{Introduction}
\label{sec:introduction}

\begin{figure}[ht]
    \centering
    \includegraphics[width=\linewidth]{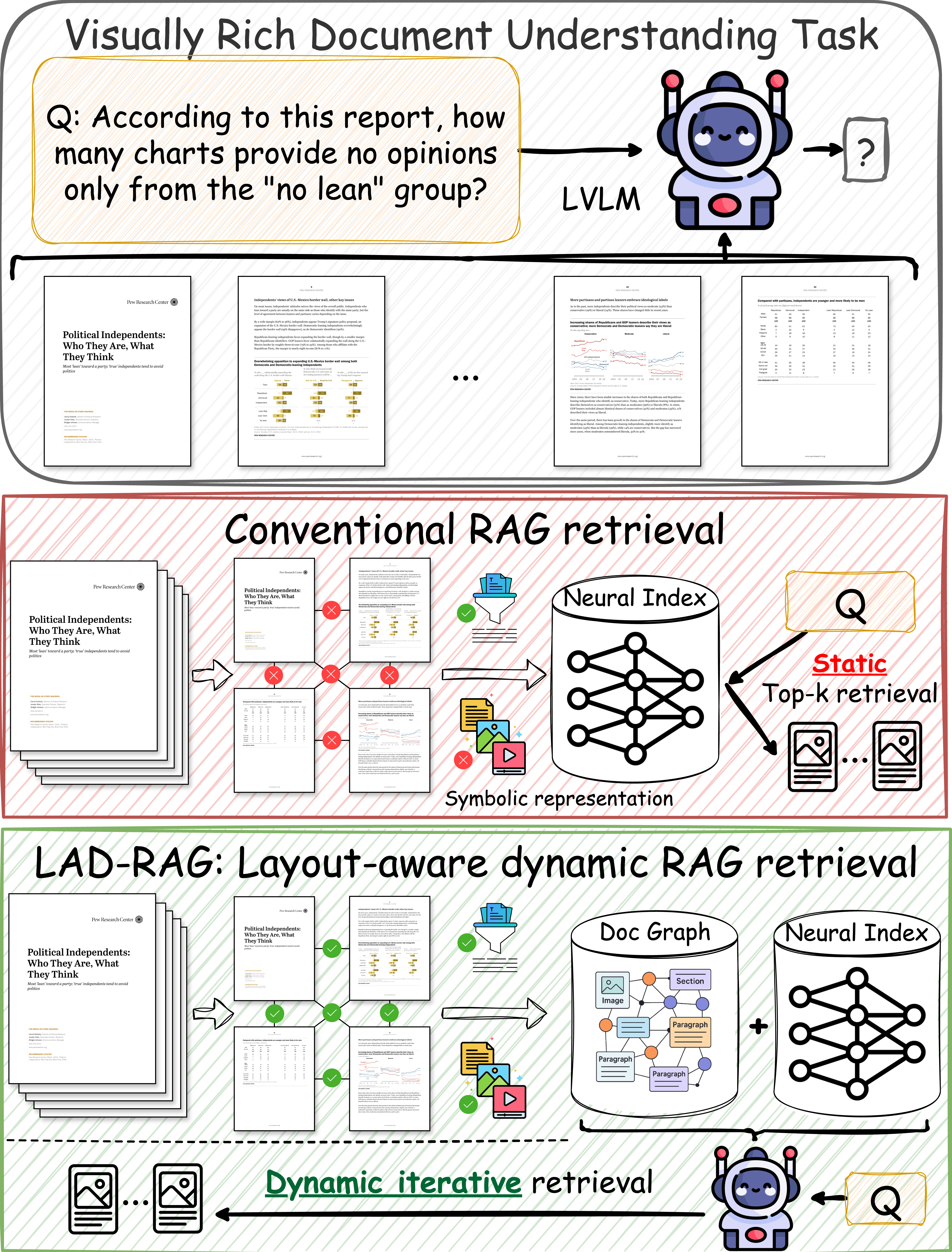}
    \caption{
    LAD-RAG addresses three key limitations of conventional RAGs in VRDs by (1) introducing a symbolic document graph to capture layout and cross-page structure, (2) integrating symbolic and neural indices to preserve structural and semantic signals, and (3) leveraging an LLM agent for dynamic, query-adaptive retrieval beyond static top‑$k$ methods.
    }
    \label{fig:main-figure}
\end{figure}

Performing NLP tasks such as question answering, summarization, and data extraction over visually rich documents (VRDs; \citealp{xu2020layoutlmv2}) requires processing text, figures, charts, and reasoning over layout (e.g., document structure, reading order, and visual grouping; \citealp{masry2022chartqa,wu2024chartinsights,xu2023chartbench,kahou2017figureqa,luo2024layoutllm}). Modern multimodal models (e.g., GPT4o; \citealp{hurst2024gpt}, InternVL; \citealp{chen2024internvl}, Qwen2.5VL; \citealp{bai2025qwen2}) are capable of processing such inputs \citep{wang2025mmlongbench}, but their effectiveness is limited when documents exceed the model's context window \citep{yin2024survey}. Plus, performance often degrades as input grows longer, even when the entire document fits within the context \citep{zhou2024rethinking,wang2024multimodal}, as relevant signals become diluted by noise, leading to incomplete or incorrect answers \citep{sharma2024losing,zhang2023siren}.

To overcome these issues, retrieval-augmented generation (RAG; \citealp{lewis2020retrieval,shi2023replug}) frameworks index document chunks during ingestion and retrieve a subset of relevant chunks at inference time to help generate and ground the answer in a given document \citep{gao2023retrieval,chen2022murag}. However, as shown in \Cref{fig:main-figure}, conventional RAG approaches rely mostly on dense text and image encoders \citep{khattab2020colbert,faysse2024colpali,karpukhin2020dense}, 
and treat document segments as a linear sequence of isolated units \citep{duarte2024lumberchunker,plonka2025comparative,han2024retrieval}, neglecting document structure \citep{wang2025document} and inter-page relationships. This results in three major limitations:

\begin{enumerate}[leftmargin=*]
    \item \textbf{Loss of layout and structural context}. Ignoring layout-driven hierarchies and cross-page continuity can lead to incomplete evidence retrieval. For example, in response to \textit{``How many businesses are shown as examples of transformation by big data?''}, conventional retrievers may only return the title slide, missing subsequent pages listing the examples. This happens as the examples are connected with the title slide through the document's structure and layout.

    \item \textbf{Over-reliance on embeddings}. Retrieval struggles with queries depending on symbolic or structural cues (e.g., references to charts, page numbers, or table sources). Consider the query \textit{``How many charts and tables in this report are sourced from Annual totals of Pew Research Center survey data?''} Answering this requires aggregating non-contiguous, yet structurally related, figures and captions that are not explicitly captured in semantic embeddings.

    \item \textbf{Static top-$k$ retrieval}. Retrieval depth is agnostic to question or document complexity, often leading to retrieving too much or too little evidence. Consider the differing scope needed for various questions: \textit{How many organizations are introduced in detail?''} need only three pages, but \textit{How many distinct Netherlands location images are used as examples?''} require twelve.

\end{enumerate}

In short, existing RAG pipelines lack a holistic document representation at ingestion time, which hinders the retrieval of a complete set of evidence, particularly when relevant content is structurally dispersed across the document \citep{ma2024mmlongbench,deng2024longdocurl}. To address this, we introduce \textbf{LAD-RAG}, a layout-aware dynamic RAG. As illustrated in \Cref{fig:main-figure}, LAD-RAG enhances conventional neural indexing with a symbolic document graph built at ingestion. Nodes in this graph encode explicit symbolic elements (e.g., headers, figures, tables), while edges capture structural and layout-based relationships (e.g., section boundaries, figure–caption links, and cross-page dependencies). This design supports both fine-grained retrieval over individual nodes and higher-level retrieval over structurally grouped elements (e.g., all components of a section), enabling multiple complementary retrieval pathways.

Given a query at inference time, to effectively leverage the indexed information, a language model agent accesses both the neural index and the document graph to determine an appropriate retrieval strategy: neural, graph-based, or hybrid, and iteratively interacts with both indices to retrieve a complete set of evidence. Critically, because the document graph encodes both local and global structural relationships, including layout-based neighborhoods and higher-order patterns such as community partitions, LAD-RAG supports contextualization of retrieved nodes into coherent and complete groups of nodes. This supports the extraction of complete and well-structured evidence sets, in contrast to the partial and fragmented retrievals typical of traditional RAG systems.

We evaluate LAD-RAG on four challenging VRD benchmarks, MMLongBench-Doc, LongDocURL, MP-DocVQA, and DUDE, with diverse layouts (slides, reports, papers) and questions requiring evidence scattered across multiple pages. LAD-RAG consistently improves both retrieval (achieving over 90\% perfect recall on average without any top‑$k$ tuning, and outperforming baseline retrievers by up to 20\% in recall at comparable noise levels) and QA accuracy, approaching the performance with ground-truth evidence, while introducing minimal inference latency.

As RAG frameworks become central to grounding LLMs in real documents, reducing hallucination and improving the reliability of their answers \citep{bechard2024reducing}, our work highlights the importance of representing the stored documents during this process in a way that more closely mirrors how humans approach the same problem: when we read a document, we form a connected mental picture of its content \citep{saux2021building}, and depending on the question and how challenging it is, we naturally look back at different parts of the document and rely on different amounts of supporting evidence. Conventional RAGs often miss this coherence, treating documents as a set of separate pieces and retrieving information in a fixed, question-agnostic way. LAD-RAG moves closer to how a person would reason through the task: it keeps a more unified view of the document, preserves the links between related sections and elements, and adjusts how much information it pulls in based on what the question actually needs, leading to more complete and grounded answers.

\section{LAD-RAG framework}

\begin{figure*}[ht]
    \centering
    \includegraphics[width=1.0\linewidth]{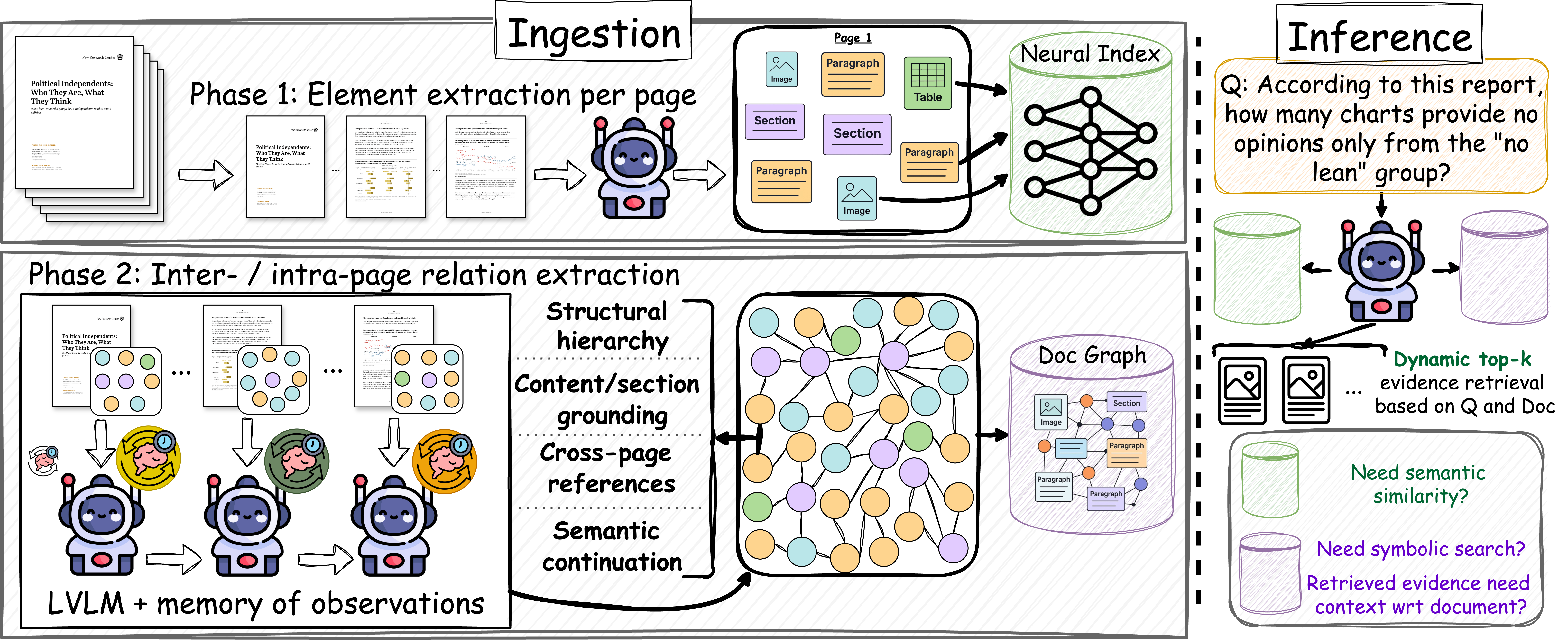}
    \caption{
    LAD-RAG framework: During ingestion, an LVLM extracts elements from each page and encodes them into an index, while also constructing a document graph that captures inter- and intra-page relationships for symbolic search. During inference, an agent interprets the question and iteratively interacts with both the neural index and the document graph to retrieve relevant evidence, enabling question-aware, structure-aware retrieval.
    }
    \label{fig:framework}
\end{figure*}

Given a document ($d$) that is the target of a question ($q$), the goal of our LAD-RAG framework is to construct a \emph{holistic, contextualized} understanding of $d$'s content at ingestion time to support more complete and accurate retrieval at inference time. Achieving this requires organizing the framework into two corresponding phases, illustrated in \Cref{fig:framework}: (i) preparing and storing rich document information both neurally and symbolically during \textbf{ingestion} (\Cref{subsec:framework-ingestion-time}), and (ii) using that information strategically during \textbf{inference} (\Cref{subsec:framework-inference-time}).

\subsection{Ingestion}
\label{subsec:framework-ingestion-time}

At ingestion time, we aim to comprehensively parse the document $d$ and store its content in both neural and symbolic forms. Inspired by how human readers sequentially build mental models by reading page by page while tracking cross-references and higher-order themes \citep{saux2021building}, we simulate a similar process, as demonstrated in \Cref{fig:framework}.

To support flexible understanding across diverse document layouts, we use GPT-4o \citep{hurst2024gpt}, a powerful large vision–language model (LVLM), to process each page in sequence (see \Cref{app:implementation-details} for details). As the LVLM parses each page, it extracts all visible elements and generates self-contained descriptions for each (which will later form the \textbf{nodes} of the document graph).

Beyond page-level processing, we maintain a running \textbf{memory ($M$)}, akin to a human reader's ongoing understanding of a document \citep{saux2021building}. This memory accumulates key high-level information, e.g., section structure, entity mentions, and thematic progressions, across pages. As each new page is processed, we connect its elements with relevant parts of $M$ to build inter-page relationships, as the \textbf{edges} in the document graph.

Once the model has completed a full pass over the document, we construct a full \textbf{document graph ($G$)} containing both intra- and inter-page structure. This graph is stored in two complementary forms: the full symbolic document graph object and a neural index over its nodes, each enabling different modes of downstream retrieval.

\paragraph{Document graph nodes.}
Each node corresponds to a localized element on a page, such as a paragraph, figure, table, section title, or footnote. For every element, we extract
\begin{itemize*}[label={}]
    \item Layout position on the page, 
    \item Element type (e.g., figure, paragraph, section header, etc.), 
    \item Displayed content, including extractive text or semantic,  captions, 
    \item Self-contained summary that enables standalone interpretation, and 
    \item Visual attributes (e.g., font, color, size)
\end{itemize*}

This node-level representation ensures that every element can be retrieved and interpreted in isolation \citep{bhattacharyya2025information}, if necessary, while also enabling structured indexing.

\paragraph{Document graph edges.}
Edges connect nodes based on symbolic, layout, or semantic relationships, including:
\begin{itemize*}[label={}]
    \item Reference relationships, e.g., a paragraph referring to a figure, or a footnote referring to a section, and 
    \item Layout/structural relationships, e.g., elements belonging to the same section or representing cross-page continuations.
\end{itemize*}

Constructing these edges requires a higher-level understanding of the document's structure. This is enabled by the running memory ($M$), which tracks the evolving context, including the current section hierarchy, the active entities under discussion, and unresolved references (e.g., placeholders for elements explained later). These contextual signals help disambiguate relationships that are not evident from the current page alone.

\paragraph{Neural–symbolic indexing.}
The final output of ingestion is stored in two complementary representations:
\begin{enumerate*}
    \item Symbolic (graph) index ($G$): A document graph with structured node/edge properties (e.g., content type, location, visual attributes) and explicit local and global relationships for graph-based retrieval, capturing a set of nodes part of a community captured by the graph structure.
    \item Neural index ($E$): A vector-based index over the self-contained summaries of all nodes, enabling semantic similarity search.
\end{enumerate*}

This dual representation preserves both the semantic richness of neural models and the explicit structure captured in the document's layout, enabling retrieval mechanisms that would otherwise be inaccessible through embedding-based approaches alone, which can summarize and abstract away from the input content, making nuances and details explicitly inaccessible.

\subsection{Inference}
\label{subsec:framework-inference-time}

At inference time, given a question $q$, its symbolic graph $G$, and its neural index $E$, the goal is to retrieve a complete and contextually appropriate set of evidence from $G$ and/or $E$. This process is both query-dependent (some questions can be answered via embedding similarity, while others require symbolic reasoning over layout or structure) and interactive, as the scope and difficulty of the query may require varying amounts of evidence.

To accommodate these needs, LAD-RAG employs an LLM agent \citep[GPT-4o,][]{hurst2024gpt} that iteratively interacts with both indices to retrieve the full set of evidence required to answer the question (see \Cref{fig:framework}).

The LLM agent is equipped with tool interfaces with explicit function signatures to access and operate over the indices. Given the query, the agent generates a high-level plan (e.g., selecting between semantic, symbolic, or hybrid retrieval modes), issues tool calls accordingly, and iteratively refines the evidence set through a conversational loop. This loop terminates when one of the following conditions is met: (i) nearing the model's context window, (ii) reaching a maximum number of steps, or (iii) the agent determines that sufficient evidence has been gathered. See \Cref{app:implementation-details} for details.

The following tools are exposed to the agent:
\begin{enumerate*}
\item \textbf{NeuroSemanticSearch(query):} Retrieves evidence based on embedding similarity from the neural index, using a query composed by the agent based on the given question $q$.
\item \textbf{SymbolicGraphQuery(query\_statement):} Performs structured queries over the symbolic document graph (e.g., filtering by element type, section, or page). The agent is instructed on the graph representation and must generate query statements to interact with the document graph object and extract relevant nodes based on their properties or structural position.
\item \textbf{Contextualize(node):} Expands a given node into a broader evidence set based on its structural proximity within the graph. This expansion leverages both local neighborhoods and higher-order relationships, using Louvain community detection \citep{blondel2008fast} to surface coherent clusters of contextually related nodes. Additional implementation details are provided in \Cref{app:implementation-details}.
\end{enumerate*}

Together, these tools enable the system to flexibly retrieve evidence tailored to the specific query, document, and reasoning complexity, going beyond fixed top-$k$ retrieval to support fully contextualized and adaptive evidence selection.

\section{Experimental Setup}

\subsection{Datasets}

To evaluate LAD-RAG, we conduct experiments on four diverse benchmarks for visually rich document understanding: MMLongBench-Doc \citep{ma2024mmlongbench}, LongDocURL \citep{deng2024longdocurl}, DUDE \citep{van2023document}, and MP-DocVQA \citep{tito2023hierarchical}. These datasets cover a wide range of domains and question types, including those grounded in high-level document content and those dependent on more localized visual elements such as figures, references, sections, and tables. They also vary in the number of pages required to answer questions, making them well-suited for evaluating retrieval completeness as well as the downstream question answering performance of LAD-RAG. See \Cref{app:dataset-details} for details.

\subsection{Baselines}

\paragraph{Retrieval.}  
We compare LAD-RAG's evidence retrieval performance with both text-based and image-based retrievers. The text-based baselines include E5-large-v2 \citep{wang2022text}, BGE-large-en \citep{xiao2024c}, and BM25 \citep{lu2024bm25s}, while the image-based baseline is ColPali \citep{faysse2024colpali}, which is closely aligned with the retrieval backbone used in M3DocRAG \citep{cho2024m3docrag}, one of the strongest multimodal RAG systems over VRDs. These baselines follow the standard RAG setup, operating over summaries of all extracted page elements, retrieving the top-$k$ elements based on similarity to the query, at inference time. We evaluate performance across $k$ values up to the point of perfect recall. We additionally compare against RAPTOR \citep{sarthi2024raptor}, a hierarchical RAG baseline that groups semantically related document elements into aggregate nodes, each summarizing its constituents. At inference time, retrieving an aggregate node expands to include all its constituent nodes down to the leaf level, enabling retrieval across multiple levels of granularity.


\paragraph{Question answering.}  
To assess the downstream QA performance of our system, we pair the retrieved evidence with four LVLMs: Phi-3.5-Vision-4B \citep{abdin2024phi}, Pixtral-12B-2409 \citep{agrawal2024pixtral}, InternVL2-8B \citep{chen2024internvl}, and GPT-4o \citep{hurst2024gpt}.\footnote{We also tested LLaMA 3.2 Vision but filtered it out after it failed multi-image perception tasks in preliminary testing.} For each model, we use
deterministic greedy decoding for answer generation and compare QA accuracy under several retrieval settings:
\begin{itemize}[leftmargin=*]
    \item Using evidence retrieved by LAD-RAG.
    \item Using the best-performing baseline retriever at fixed retrieval sizes: $k=5$ and $k=10$.
    \item Using the same best-performing baseline retriever, matching the number of retrieved items with LAD-RAG (to control for retrieval budget).
    \item Using ground-truth evidence pages (oracle retrieval) as an upper bound.
\end{itemize}
To illustrate the value of retrieval for QA in VRD, we also compare against models that receive the full document as input, i.e., mPLUG-DocOwl v1.5-8B \citep{hu2024mplug}, Idefics2-8B \citep{laurenccon2024matters}, and MiniCPM-Llama3-V2.5-8B \citep{yao2024minicpm}, highlighting the effectiveness of retrieval-augmented approaches in reducing noise while maintaining high accuracy.

\subsection{Evaluations \& Metrics}
\label{subsec:evaluation-metrics}

\paragraph{Retrieval.}  
Let a question $q$ be associated with a document $d$, and let the ground-truth set of evidence pages be denoted by $P = \{p_1, p_2, \ldots, p_n\}$. Given a retriever that returns a set of pages $\hat{P}$, our goal is to assess how well $\hat{P}$ matches $P$, both in terms of completeness and precision. We report two key retrieval metrics:

\begin{itemize}[leftmargin=*]
    \item \textbf{Perfect Recall (PR):} A binary metric defined as:
    \[
    \text{PR} =
    \begin{cases}
    1 & \text{if } P \subseteq \hat{P} \\
    0 & \text{otherwise}
    \end{cases}
    \]
    It indicates whether the retriever has retrieved \emph{all} ground-truth pages. This is crucial for multi-page questions, where missing even a single evidence page may result in incorrect answers.

    \item \textbf{Irrelevant Pages Ratio (IPR):} The proportion of retrieved pages that are \emph{not} in the gold set:
    \[
    \text{IPR} = \frac{|\hat{P} \setminus P|}{|\hat{P}|}
    \]
    capturing the noise the retriever introduces, with lower values indicating a more targeted retrieval.
\end{itemize}

An ideal retriever achieves high perfect recall while minimizing the irrelevant pages ratio, ensuring that all necessary evidence is retrieved without excessive inclusion of irrelevant content.

\paragraph{Question answering.}  
Following the evaluation setup in MMLongBench~\citep{ma2024mmlongbench} and LongDocURL~\citep{deng2024longdocurl}, we use GPT-4o~\citep{hurst2024gpt} to extract concise answers from retrieved content and apply a rule-based comparison to assign binary correctness, based on which we report \textbf{accuracy}. To validate this automated evaluation, we conducted a human evaluation study confirming strong alignment between human judgments and our automated judge (see \Cref{app:human-eval}).


\section{Results}

\subsection{Retrieval Effectiveness of LAD-RAG}
\label{subsec:results-retrieval}

\begin{figure*}[ht]
    \centering
    \includegraphics[width=\linewidth]{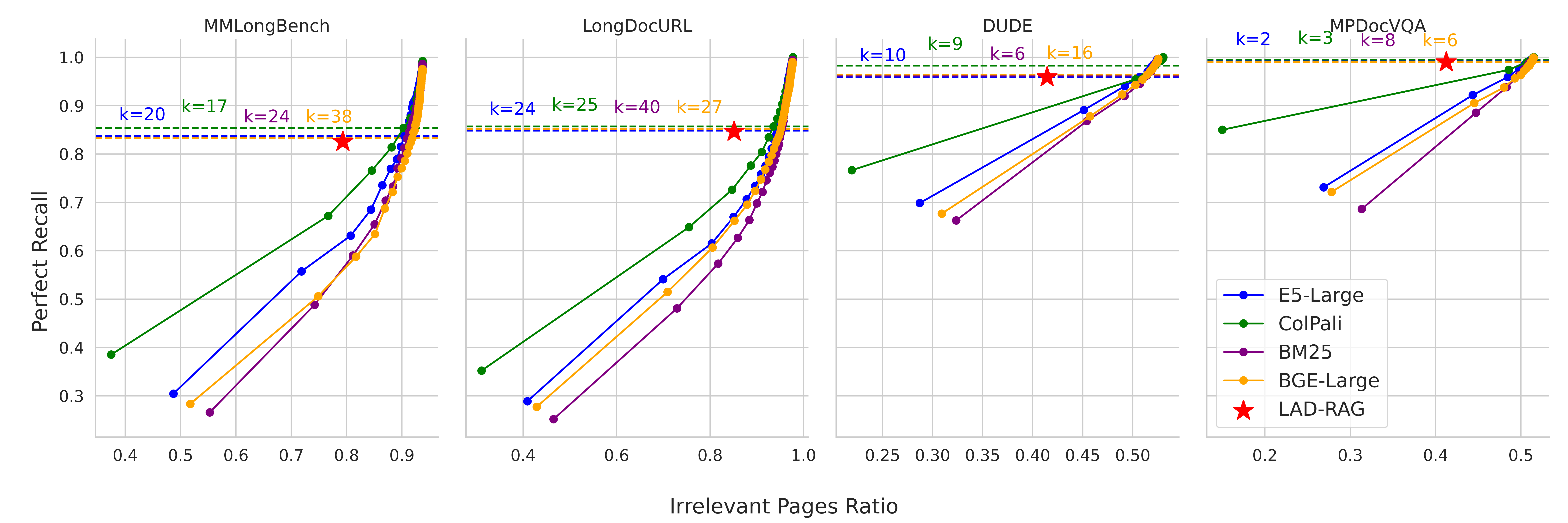}
    \caption{Retrieval performance of LAD-RAG compared to baseline retrievers across varying top-$k$ settings. Baselines retrieve from $k=1$ up to the point of perfect recall. Dotted horizontal lines indicate the number of retrieved pages each baseline requires to match the recall achieved by LAD-RAG without any top-$k$ tuning.}
    \label{fig:main-retrieval-all}
\end{figure*}

We evaluate the retrieval performance of LAD-RAG across four challenging VRD benchmarks and baseline methods in \Cref{fig:main-retrieval-all}. Notably, LAD-RAG achieves over 90\% perfect recall on average across datasets \emph{without any top-$k$ tuning}. Despite this, it retrieves significantly fewer irrelevant pages compared to existing baselines. At equivalent irrelevance rates, LAD-RAG outperforms other baselines by approximately 20\% on MMLongBench-Doc, 15\% on LongDocURL, and 10\% on both DUDE and MP-DocVQA, in the perfect recall rate.

In contrast, baseline retrievers require large top‑$k$ values to match our recall rate: on average, $k=22$ for MMLongBench-Doc, $k=27$ for LongDocURL, $k=10$ for DUDE, and $k=5$ for MP-DocVQA. These findings reveal a critical mismatch between common retrieval practices (where $k$ is typically capped at low numbers; \citealp{cho2024m3docrag,han2025mdocagent}) and the actual evidence volume required to answer multi-page questions. This also partially explains the low QA performance (30–40\% accuracy) reported by state-of-the-art models on complex VRD benchmarks such as MMLongBench-Doc \citep{ma2024mmlongbench} and LongDocURL \citep{deng2024longdocurl}.

Among the baselines, RAPTOR \citep{sarthi2024raptor}, which constructs hierarchical aggregations of semantically related content, must retrieve substantially more irrelevant pages to match LAD-RAG's recall level. Similarly, ColPali \citep{faysse2024colpali}, which serves as the retrieval backbone of M3DocRAG \citep{cho2024m3docrag}, one of the strongest multimodal RAG systems for VRD, also accumulates significantly more noise before reaching comparable recall. Together, these results indicate that neither semantic hierarchies nor vision-based retrieval alone are sufficient substitutes for the symbolic layout structure that LAD-RAG encodes, and that more selective retrieval requires explicit modeling of how document elements relate structurally, not just semantically or visually.

The advantage of LAD-RAG is even more pronounced in questions requiring evidence from multiple pages. LAD-RAG consistently retrieves more complete multi-page evidence sets than all baselines (see \Cref{app:additional-analysis-on-retrieval}). This reflects its ability to capture distributed information through structured layout modeling and dynamic interaction with the symbolic and the neural indices. For a deeper understanding of these benefits, we present detailed case studies in \Cref{case-study}.

\subsection{Ablation Study of Retrieval Components}

To understand the contribution of individual components in LAD-RAG, we conduct an ablation study on MMLongBench and LongDocURL, the most challenging benchmarks, comparing overall retrieval performance (measured as the ratio of perfect recall to irrelevant page retrievals, with higher values indicating better recall and lower noise) across LAD-RAG variants. We compare the full LAD-RAG (with both \texttt{Contextualize} and \texttt{GraphQuery}) against versions that disable one or both components. As shown in \Cref{tab:retrieval-composite}, removing either contextualization (C) or symbolic graph querying (G) results in noticeable performance drops: at similar noise levels, recall decreases by an average of 4\% without contextualization and 10\% without graph querying. Notably, the variant without graph or contextualization (LAD-RAG w/o C \& G), which is a retrieval agent interacting only with the neural index, still outperforms conventional baselines. This is analogous in spirit to the dynamic retrieval pipeline proposed by SimpleDoc \citep{jain2025simpledoc}, and the fact that our full method outperforms this variant confirms that the symbolic graph and graph-based retrieval provide gains beyond dynamic neural retrieval alone.

We also observe that RAPTOR \citep{sarthi2024raptor}, which provides hierarchical aggregations of semantically related content, performs similarly to our LAD-RAG variant without graph querying (LAD-RAG w/o G), suggesting that RAPTOR's hierarchical groupings provide benefits analogous to our contextualization component. However, it does not match the additional gains provided by LAD-RAG's symbolic structure and agent-driven retrieval, confirming that semantic hierarchies alone are insufficient to fully capture layout-driven and cross-page dependencies. Together, these results show that while symbolic and graph-based mechanisms are essential, the interactive nature of the retriever also contributes to assembling a complete evidence set.

\begin{table}[ht]
\centering
\resizebox{\columnwidth}{!}{%
\begin{tabular}{lcc}
\toprule
\textbf{Method} & \textbf{MMLongBench} & \textbf{LongDocURL} \\
\midrule
LAD-RAG                   & 0.979 & 0.895 \\
LAD-RAG w/o C             & 0.957 & 0.819 \\
LAD-RAG w/o G             & 0.856 & 0.809 \\
LAD-RAG w/o C \& G     & 0.840 & 0.774 \\
RAPTOR & 0.877 & 0.853 \\
ColPali                & 0.831 & 0.791 \\
BM25                   & 0.728 & 0.762 \\
E5-Large            & 0.791 & 0.769 \\
BGE-Large           & 0.743 & 0.704 \\
\bottomrule
\end{tabular}
}
\caption{Retrieval performance of LAD-RAG variants, measured as the ratio of perfect recall to irrelevant page retrievals (higher is better). \texttt{C} and \texttt{G} denote \texttt{Contextualize} and \texttt{GraphQuery} operations.}
\label{tab:retrieval-composite}
\end{table}

\begin{table*}[t]
\centering
\resizebox{\textwidth}{!}{%
\begin{tabular}{llcccccccccc}
\toprule
Model & Retrieval & \multicolumn{3}{c}{\textbf{MMLongBench-Doc}} & \multicolumn{3}{c}{\textbf{LongDocURL}} & \multicolumn{3}{c}{\textbf{DUDE}} & \multicolumn{1}{c}{\textbf{MP-DocVQA}} \\
\cmidrule(lr){3-5}\cmidrule(lr){6-8}\cmidrule(lr){9-11}\cmidrule(lr){12-12}
 &  & all & single & multi & all & single & multi & all & single & multi & all \\
\midrule
mPLUG-DocOwl 1.5-8B & All-Pages & 0.069 & 0.074 & 0.064 & 0.031 & 0.039 & 0.024 & 0.150 & 0.188 & 0.116 & 0.150 \\
Idefics2-8B & All-Pages & 0.070 & 0.077 & 0.072 & 0.045 & 0.054 & 0.038 & 0.170 & 0.205 & 0.130 & 0.170 \\
MiniCPM-Llama3-V2.5-8B & All-Pages & 0.085 & 0.095 & 0.095 & 0.060 & 0.066 & 0.053 & 0.190 & 0.210 & 0.150 & 0.190 \\
\midrule
\multirow{5}{*}{InternVL2-8B} & Ground-Truth & 0.399 & 0.506 & 0.250 & 0.629 & 0.684 & 0.580 & 0.641 & 0.697 & 0.574 & 0.848 \\
 & retrieving@5 & 0.287 & 0.372 & 0.164 & 0.443 & 0.484 & 0.404 & 0.560 & 0.564 & 0.367 & 0.737 \\
 & retrieving@10 & 0.319 & 0.395 & 0.208 & 0.457 & 0.499 & 0.420 & 0.576 & 0.591 & 0.350 & 0.759 \\
 & topk-adjusted & 0.304 & 0.365 & 0.212 & 0.468 & 0.515 & 0.411 & 0.589 & 0.609 & 0.416 & 0.773 \\
 & \textbf{LAD-RAG} & \textbf{0.448} & \textbf{0.495} & \textbf{0.242} & \textbf{0.477} & \textbf{0.518} & \textbf{0.438} & \textbf{0.630} & \textbf{0.669} & \textbf{0.533} & \textbf{0.792} \\
\addlinespace[2pt]
\midrule
\multirow{4}{*}{Pixtral-12B-2409} & Ground-Truth & 0.498 & 0.537 & 0.343 & 0.634 & 0.668 & 0.603 & 0.602 & 0.636 & 0.467 & 0.839 \\
 & retrieving@5 & 0.389 & 0.395 & 0.200 & 0.430 & 0.458 & 0.401 & 0.514 & 0.512 & 0.417 & 0.663 \\
 & retrieving@10 & 0.345 & 0.327 & 0.178 & 0.427 & 0.457 & 0.400 & 0.529 & 0.560 & 0.417 & 0.655 \\
 & topk-adjusted & 0.383 & 0.394 & 0.213 & 0.455 & 0.488 & 0.423 & 0.535 & 0.574 & 0.426 & 0.675 \\
 & \textbf{LAD-RAG} & \textbf{0.415} & \textbf{0.462} & \textbf{0.261} & \textbf{0.507} & \textbf{0.551} & \textbf{0.468} & \textbf{0.545} & \textbf{0.578} & \textbf{0.433} & \textbf{0.704} \\
\addlinespace[2pt]
\midrule
\multirow{5}{*}{Phi-3.5-Vision-4B} & Ground-Truth & 0.383 & 0.492 & 0.227 & 0.631 & 0.659 & 0.606 & 0.578 & 0.638 & 0.446 & 0.810 \\
 & retrieving@5 & 0.315 & 0.399 & 0.192 & 0.476 & 0.485 & 0.468 & 0.562 & 0.615 & 0.422 & 0.719 \\
 & retrieving@10 & 0.302 & 0.381 & 0.189 & 0.479 & 0.483 & 0.476 & 0.576 & 0.643 & 0.450 & 0.737 \\
 & topk-adjusted & 0.336 & 0.400 & 0.199 & 0.469 & 0.484 & 0.455 & 0.582 & 0.656 & 0.466 & 0.737 \\
 & \textbf{LAD-RAG} & \textbf{0.391} & \textbf{0.414} & \textbf{0.202} & \textbf{0.489} & \textbf{0.493} & \textbf{0.476} & \textbf{0.596} & \textbf{0.656} & \textbf{0.467} & \textbf{0.754} \\
\addlinespace[2pt]
\midrule
\multirow{5}{*}{GPT-4o-200b-128} & Ground-Truth & 0.696 & 0.693 & 0.565 & 0.714 & 0.746 & 0.686 & 0.807 & 0.830 & 0.633 & 0.895 \\
 & retrieving@5 & 0.575 & 0.607 & 0.303 & 0.590 & 0.684 & 0.510 &  0.707 & 0.732 & 0.534 & 0.825 \\
 & retrieving@10 & 0.610 & 0.637 & 0.372 & 0.622 & 0.702 & 0.552 & 0.706 & 0.732 & 0.535 & 0.819 \\
 & topk-adjusted & 0.593 & 0.629 & 0.409 & 0.652 & 0.709 & \textbf{0.600} & 0.720 & \textbf{0.750} & 0.541 & \textbf{0.833} \\
 & \textbf{LAD-RAG} & \textbf{0.625} & \textbf{0.676} & \textbf{0.450} & \textbf{0.659} & \textbf{0.724} & 0.599 & \textbf{0.725} & 0.746 & \textbf{0.545} & 0.829 \\
\bottomrule
\end{tabular}
}
\caption{Accuracy scores per model across retrieval types (topk-adjusted: evidence with the same number of retrieved pages as LAD-RAG) and top-$k$. Single/multi refer to questions requiring evidence from one or multiple pages, respectively. Best values per model group (excluding \texttt{GT}) are shown in \textbf{bold}.}
\label{tab:extended_scores_with_topk}
\end{table*}

\subsection{LAD-RAG's End-to-End QA Gains}
\label{subsec:results-QA}

While our core contribution is retrieval, we also observe consistent gains in downstream QA accuracy across all models and benchmarks when using LAD-RAG. Results are shown in \Cref{tab:extended_scores_with_topk}. Across both smaller models (e.g., InternVL2‑8B, Phi-3.5-Vision) and larger models (e.g., GPT‑4o‑200B), LAD-RAG consistently improves QA accuracy over all retrieval baselines and the ones that access the entire document without retrieval. It outperforms both fixed top‑$k$ retrieval ($k=5,10$) and the topk‑adjusted setting, where baselines retrieve the same number of pages as our system. This shows that our retriever not only supplies more relevant evidence but also introduces substantially less noise, directly translating to higher downstream accuracy.

Focusing on the multi-page questions that constitute the more challenging subset of examples, LAD-RAG yields consistent gains across all four benchmarks, averaging 4 and up to 18 percentage points over top‑$k$ baselines, and 3 and up to 11 points over top‑$k$-adjusted retrieval (see \Cref{app:additional-analysis-qa} for similar trends across all question and document types). This indicates that LAD-RAG's improvements stem not just from retrieving more pages, but from retrieving more focused and complete evidence.

Notably, LAD-RAG's end-to-end performance approaches that of using ground-truth evidence across all datasets, narrowing the gap to within 5–8 points, especially for challenging benchmarks like MMLongBench and LongDocURL. This contextualizes both the upper bound of current models and the significance of improved retrieval. It is worth noting that ground-truth evidence does not yield perfect QA accuracy, and in a small number of cases LAD-RAG even surpasses it. This occurs when models benefit from more contextualized or expanded evidence: even if some retrieved content is not exactly annotated as evidence for the question, it can provide richer grounding that aids reasoning, whereas the succinct annotated spans alone may not always be sufficient for the model to arrive at the correct answer. More broadly, this reflects a complementary bottleneck in current LVLMs: even with near-perfect evidence, models may still fail to fully utilize it, suggesting that improvements to downstream reasoning remain an important and orthogonal direction for future work.

Together, these results confirm that LAD-RAG improves end-to-end QA, particularly in multi-page settings where conventional methods fall short. By assembling more focused and relevant evidence, LAD-RAG boosts answer accuracy in a way that generalizes across models and datasets.

\begin{figure}[t]
    \centering
    \includegraphics[width=1\linewidth]{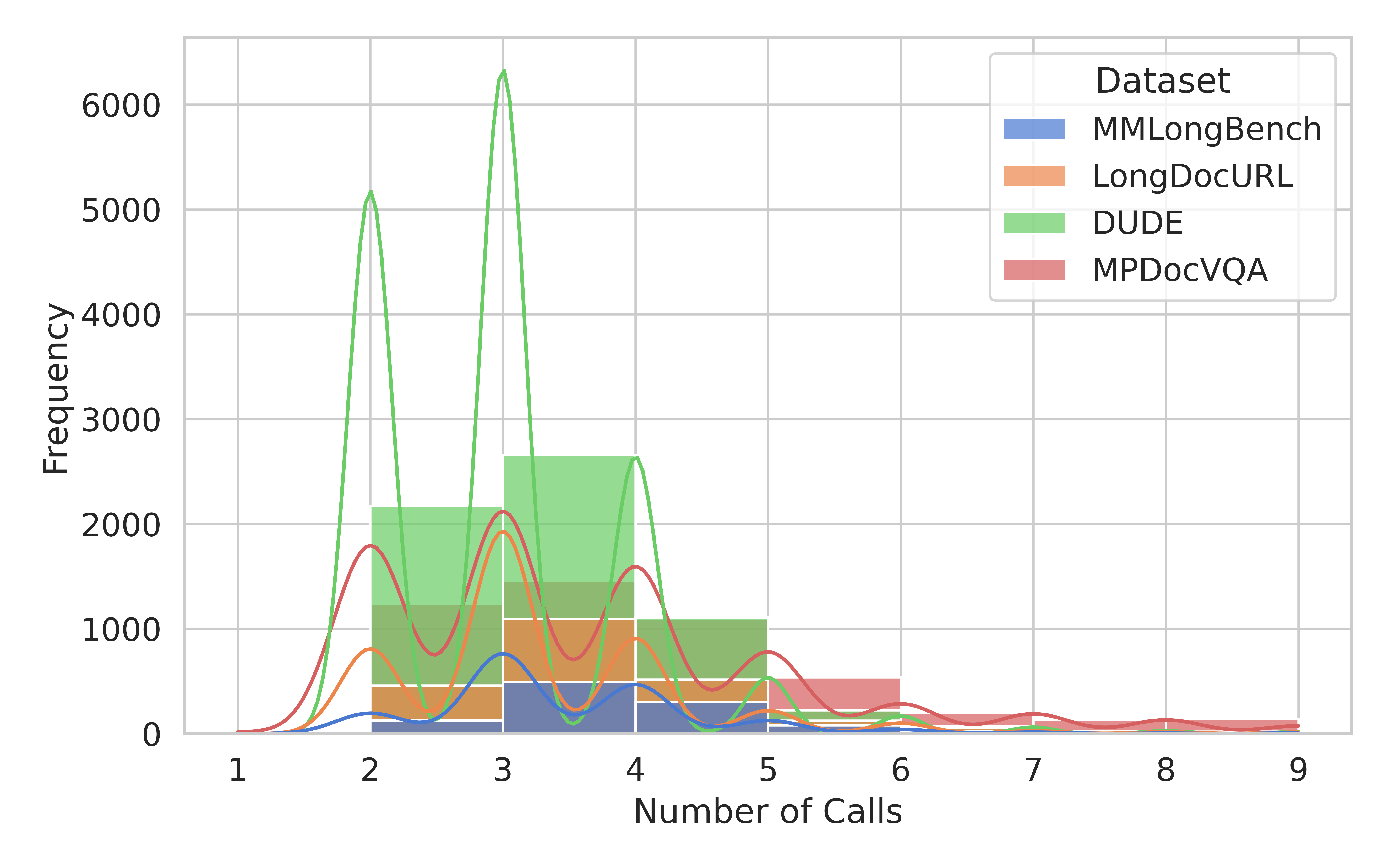}
    \caption{Distribution of the number of LLM calls per query by LAD-RAG.}
    \label{fig:llm_calls}
\end{figure}

\subsection{Latency Analysis}
\label{subsec:results-latency}

Analyzing the latency of LAD-RAG, we find that it introduces minimal overhead during inference. Graph construction is performed once, offline during ingestion, and does not affect inference-time latency. During inference, our agent-based retriever typically issues 2–5 LLM calls (\Cref{fig:llm_calls}), with over 97\% of these generating fewer than 100 tokens on average (\Cref{app:more-latency}). These tokens serve as retrieval queries executed over a pre-built symbolic graph and semantic index, both incurring negligible runtime cost. This contrasts with systems such as MoLoRAG \citep{wu2025molorag} and FRAG \citep{huang2025frag}, which expand or select evidence through expensive one-by-one inference-time scanning of document pages, incurring substantially higher retrieval costs. Altogether, this demonstrates that LAD-RAG achieves substantial QA gains with minimal added latency at inference time.

\section{Related Work}
\label{sec:related-work}
\subsection{Visually-rich Document Understanding}

With the rise of LVLMs \citep{hurst2024gpt,team2023gemini,bai2025qwen2,grattafiori2024llama,liu2023visual}, these models have been increasingly applied to document understanding tasks \citep{ma2024mmlongbench,hui2024uda,deng2024longdocurl}, especially for documents containing rich visual content \citep{kahou2017figureqa,xu2023chartbench} or spanning multiple pages \citep{hu2024mplug}. To reduce the reasoning burden on LLMs and improve response accuracy, RAG has become a dominant strategy, narrowing the context to only the most relevant parts of a document.

Several recent works have tackled the challenge of retrieving multi-modal, cross-page evidence in visually rich documents. To capture the multi-modal nature of these tasks, M3DocRAG \citep{cho2024m3docrag} and MDocAgent \citep{han2025mdocagent} combine text- and image-based retrievers with specialized reasoning agents \citep{khattab2020colbert,faysse2024colpali}, with MDocAgent relying heavily on LVLMs at inference time, incurring substantial computational cost. For modeling cross-page relationships by enriching the indices at ingestion time \citep{chen2025enrichindex}, MMRAG-DocQA \citep{gong2025mmrag} and RAPTOR \citep{sarthi2024raptor} construct hierarchical aggregations of semantically related content to improve retrieval granularity, but these groupings are fixed at ingestion and do not adapt to the specific structural or layout demands of individual queries. MoLoRAG \citep{wu2025molorag} instead expands evidence at inference time by traversing semantically similar pages, and FRAG \citep{huang2025frag} performs a full-document pass to select relevant pages or images, both effective but at the cost of expensive per-query scanning. More recently, dynamic retrieval has emerged: SimpleDoc \citep{jain2025simpledoc} iteratively refines evidence using LLM feedback, addressing the limitations of fixed top-$k$ retrieval, though without explicit modeling of layout structure or cross-page dependencies.

While these methods have advanced multi-modal retrieval, handling of cross-page dependencies still largely relies on capturing semantically similar content across pages; ingestion pipelines depend largely on embedding-based representations, losing a symbolic view of document structure; and as a result, there is no adaptability to the specific retrieval demands of different queries. LAD-RAG bridges these gaps by introducing a symbolic document graph capturing both local content and global layout structure, paired with a query-adaptive retriever that dynamically reasons over both neural and symbolic indices, achieving comprehensive evidence gathering with minimal inference-time overhead.

\subsection{RAG Frameworks with Graph Integration}

Recent RAG frameworks have enhanced retrieval through graphs. At ingestion time, several methods construct hierarchical groupings, typically based on semantic similarity or high-level layout cues like document sections, to support multi-granularity retrieval \citep{nguyen2024enhancing,wang2025document,lu2025hichunk,sarthi2024raptor,edge2024local}. However, these structures often lack flexibility for ad hoc queries that do not align with predefined boundaries. Other approaches incorporate graphs at inference time, using entity-based traversal \citep{kim2024causal,niu2024mitigating,guo2024knowledgenavigator}. Iterative graph-based retrieval has also been explored \citep{sun2023think}, along with knowledge graph prompting in multi-document QA \citep{wang2024knowledge,jiang2024reasoning,yang2025superrag}, though these approaches are typically text-only or rely on highly specialized modular pipelines that separately process textual, visual, and layout elements, making them less suited for visually rich documents with unpredictable structure.

LAD-RAG addresses the unique challenges of visually rich documents by constructing a general-purpose symbolic graph capturing both semantic and layout-based relationships, and not just localized semantic groups. Unlike modular pipeline approaches, LAD-RAG relies on a single model during ingestion with graph construction handled automatically via standard tooling, keeping deployment straightforward. At inference time, it performs dynamic, query-driven retrieval over this structure without costly inference-time traversal. By leveraging document structure to flexibly retrieve relevant node groups beyond rigid top-$k$ constraints, LAD-RAG enables comprehensive evidence gathering across pages, leading to substantial gains in QA accuracy.

\section{Conclusion}

We introduce LAD-RAG, a layout-aware dynamic RAG framework for visually rich document understanding. Unlike conventional RAG pipelines that ingest document chunks in isolation and rely solely on neural indices, LAD-RAG constructs a symbolic document graph during ingestion to capture both local semantics and global layout-driven structure. This symbolic graph is stored alongside a neural index over document elements, enabling an LLM agent at inference time to dynamically reason over and retrieve evidence based on the specific needs of each query. Across four challenging VRD benchmarks, MMLongBench-Doc, LongDocURL, DUDE, and MP-DocVQA, LAD-RAG improves retrieval completeness, achieving over 90\% perfect recall on average without any top‑$k$ tuning, and outperforming strong text- and image-based baseline retrievers by up to 20\% in recall at comparable noise levels. These gains translate to higher downstream QA accuracy, approaching oracle-level performance with minimal added latency. Our results underscore the importance of reasoning over layout and cross-page structure. LAD-RAG provides a generalizable foundation for retrieval in tasks that require contextual, multimodal understanding, with broad applicability across enterprise, legal, financial, and scientific domains.

\section{Limitations}
LAD-RAG is designed to improve retrieval completeness and precision in RAG pipelines for visually rich document understanding. While our results show consistent gains in both retrieval quality and downstream QA accuracy, we also observe that even with near-perfect evidence, current LVLMs still exhibit limitations in fully utilizing the retrieved content. Our work does not aim to enhance the reasoning capabilities of QA models themselves. Instead, our contributions are focused on improving document modeling during ingestion and leveraging that structure at inference to retrieve more relevant and complete evidence. The scope of this paper is therefore limited to retrieval improvements, not generative reasoning or answer synthesis.

Our framework relies on a powerful general-purpose LVLM to extract and structure document elements (e.g., text, tables, figures) during ingestion. Although manual inspection confirms that the extractive tasks, such as reading text from images or summarizing the content of tables and figures, were handled correctly in most cases, these models can still struggle with noisy inputs, complex layouts, or low-quality visuals. Scanned documents, such as those in DUDE \citep{van2023document} and MP-DocVQA \citep{tito2023hierarchical}, present exactly these types of challenges, as lower image quality may increase the risk of extraction errors compared to born-digital documents. MP-DocVQA consists entirely of scanned documents, while DUDE includes both scanned and born-digital content. In our experiments, manual inspection suggests that the tested LVLMs handle these scans reasonably well, with most errors stemming from downstream reasoning limitations rather than OCR or visual parsing failures; however, robustness on lower-quality inputs remains an area worth further investigation. We note that the initial parsing and extraction of document content can in many cases be performed using smaller models or traditional OCR pipelines, which may offer practical advantages in cost-sensitive settings. LVLMs are primarily needed for tasks that traditional methods struggle with, such as linking semantically and layout-related information across pages. In our preliminary experiments, we found that InternVL2-8B achieved extraction quality comparable to GPT-4o based on manual inspection of dozens of documents; we ultimately adopted GPT-4o due to its stronger instruction-following behavior and more structured output formats, which simplified downstream processing.

This also reflects a broader trade-off between using a unified model that minimizes system complexity and integrating multiple specialized tools that may boost robustness but increase engineering overhead. Compared to prior systems that rely on highly specialized modular pipelines to separately process textual, visual, and layout elements \citep{yang2025superrag}, or that lean heavily on LVLMs at inference time \citep{han2025mdocagent,huang2025frag}, LAD-RAG takes a middle path: the majority of computational load is placed in the offline ingestion phase on a single model, while graph construction is handled automatically using standard tooling and adds negligible overhead. At inference time, the agent operates over the pre-built symbolic graph using lightweight text-based reasoning, avoiding both the complexity of highly modular pipelines and the inference-time compute burden of LVLM-heavy approaches. Future work could explore further simplification of the ingestion pipeline and more lightweight variants of the framework.

\section*{Acknowledgment}  
This work was conducted during the internship of the first author at Oracle.  
We thank the Oracle internship program management team for their support, and in particular Kyu Han and Sujeeth Bharadwaj for their guidance and contributions.

\bibliography{custom}

\appendix
\renewcommand{\thesection}{\Alph{section}}  

\section{Code Availability}
\label{app:code-availability}

The implementation of LAD-RAG is currently undergoing institutional review and legal clearance prior to public release. In the meantime, we provide full transparency to support reproducibility: the paper includes detailed descriptions of all components, comprehensive hyperparameter settings, and the full set of prompt templates used throughout the pipeline (see \Cref{app:implementation-details}). The codebase will be made publicly available upon approval, and the repository link will be added in the camera-ready version. For urgent reviewer needs, the code can be privately shared upon request.

\section{Dataset Details}
\label[appsection]{app:dataset-details}

\paragraph{MMLongBench-Doc.}  
MMLongBench-Doc \citep{ma2024mmlongbench} targets long-context document understanding. It contains 1,082 expert-annotated questions over 135 lengthy PDF documents (average 47.5 pages, $\sim$21k tokens per document). Evidence comes from multimodal sources, including text, images, charts, tables, and layout structure. Notably, 33\% of questions require cross-page evidence. These properties make the benchmark especially challenging for retrieval completeness. The dataset is released under the Apache-2.0 license and is permitted for academic and research use.

\paragraph{LongDocURL.}  
LongDocURL \citep{deng2024longdocurl} integrates three task categories, understanding, reasoning, and locating, over 396 documents spanning $\sim$33k pages. It provides 2,325 high-quality QA pairs, averaging 86 pages and $\sim$43k tokens per document. Compared with earlier benchmarks \citep{ma2024mmlongbench,van2023document,tito2023hierarchical}, LongDocURL features the highest proportion of multi-page (52.9\%) and cross-element (37.1\%) questions, reflecting realistic challenges in handling heterogeneous evidence distributed across different pages and element types (e.g., paragraphs, tables, figures). The dataset is released under the Apache-2.0 license and is permitted for academic and research use.

\paragraph{DUDE.}  
The Document Understanding Dataset and Evaluation (DUDE) \citep{van2023document} is a large-scale, multi-domain benchmark built from $\sim$5k multi-page documents. It spans a wide spectrum of domains (medical, legal, technical, financial) and question types (extractive, abstractive, arithmetic, multi-hop, and non-answerable). On average, documents are 5.7 pages long with $\sim$1,831 tokens. Although both extractive and abstractive questions may require evidence from multiple pages, only the extractive subset provides explicit annotations of the evidence pages. Within this subset, questions requiring multi-page evidence constitute roughly 1\% of the total. The dataset is released under the CC BY 4.0 license and is permitted for academic and research use.

\paragraph{MP-DocVQA.}
MP-DocVQA \citep{tito2023hierarchical} extends the original DocVQA dataset \citep{mathew2021docvqa} to multi-page documents. It comprises 46k questions over 5,928 documents, totaling 47,952 pages, approximately 8 pages per document. Although the documents span multiple pages, each question is designed such that its supporting evidence is confined to a single page. As a result, the benchmark does not focus on multi-page evidence aggregation. Nevertheless, MP-DocVQA remains a valuable resource for evaluating retrieval performance in multi-page settings, particularly for assessing how well systems localize relevant information within longer documents. The dataset is released under the MIT license and is permitted for academic and research use.

\section{Case Study}
\label[appsection]{case-study}

To illustrate the advantages of LAD-RAG's dynamic and layout-aware retrieval strategy, we present two case studies. These examples demonstrate how conventional RAG methods often fall short in document-centric tasks that require structural reasoning, and how LAD-RAG's combination of symbolic document graphs, semantic search, and dynamic control enables more complete and context-aware retrieval.

\begin{figure*}[ht]
    \centering
    \includegraphics[width=\linewidth]{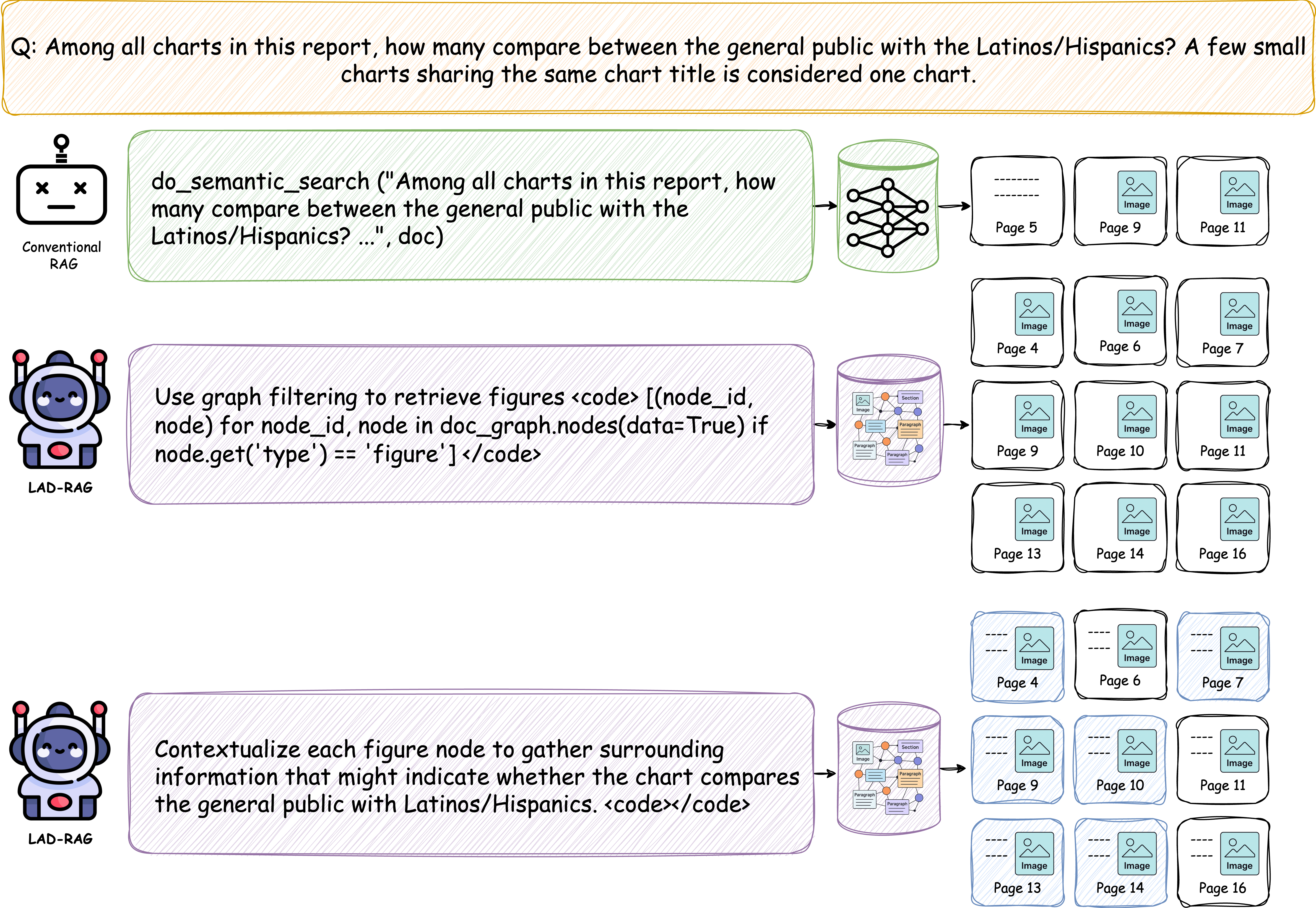}
    \caption{Case study showing LAD-RAG retrieving all charts that compare the general public with Latinos/Hispanics. While a conventional semantic retriever fails to recall many relevant charts and includes irrelevant ones, LAD-RAG dynamically opts for the symbolic retrieval. It filters for all figure nodes and then contextualizes them using the surrounding layout to determine whether they match the query. This multi-step, graph-guided process enables accurate and exhaustive evidence collection.}
    \label{fig:second-case-study}
\end{figure*}

As shown in \Cref{fig:second-case-study}, the task is to identify all charts in the report that compare the general public with Latinos/Hispanics. A conventional RAG pipeline performs semantic search using the full question string and retrieves a few charts that are semantically similar to the query. However, it misses many relevant charts, includes unrelated ones, and lacks a clear mechanism to determine how many results are needed to fully answer the question. This limitation stems from the fact that figure captions and labels are often brief, stylized, or ambiguous, making them difficult to match through semantic similarity alone.

LAD-RAG's dynamic retriever agent addresses these challenges by shifting from semantic to symbolic reasoning. It first filters the document graph to collect all nodes labeled as figures. It then contextualizes each node by examining surrounding elements such as section headers, nearby text, and related figure groups. This layout-aware process allows LAD-RAG to accurately determine whether a figure matches the query. As a result, it retrieves all relevant charts, including those distributed across pages or sharing titles. This example illustrates the advantages of dynamic control and symbolic reasoning in enabling more complete and precise retrieval for complex document-centric queries.

\begin{figure*}[ht]
    \centering
    \includegraphics[width=\linewidth]{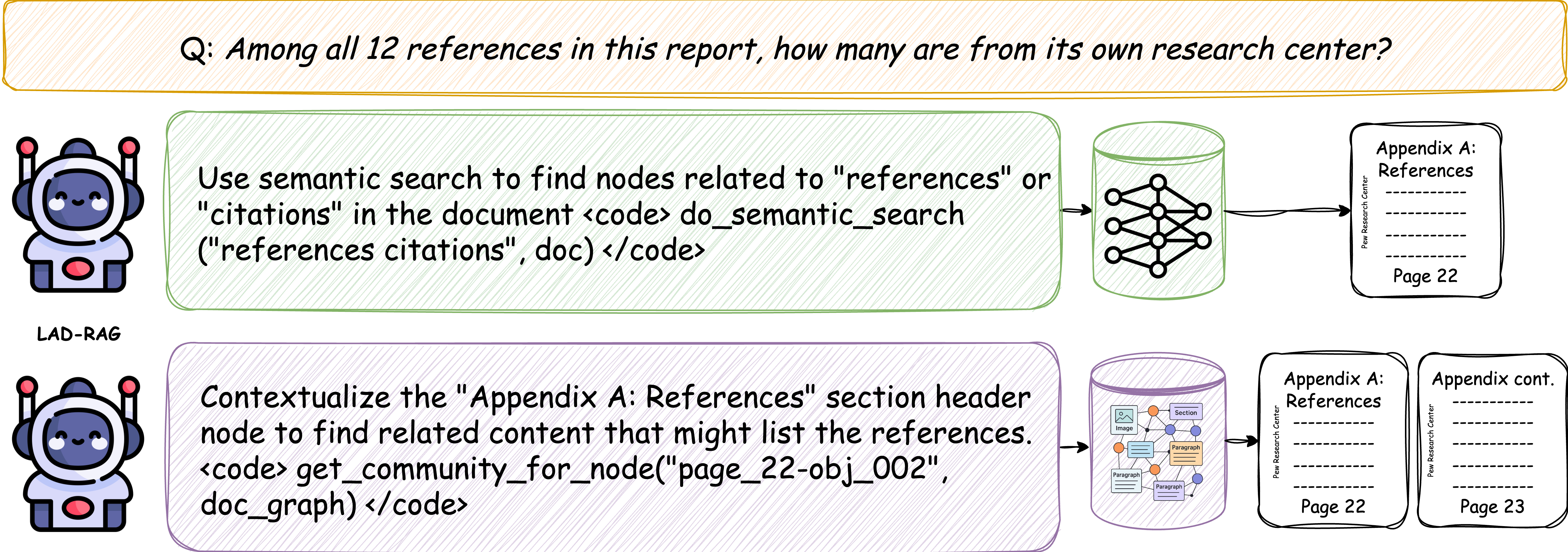}
    \caption{Case study showing LAD-RAG retrieving a multi-page reference section. While semantic search finds only the first page of references, it misses the continuation due to weak semantic overlap. LAD-RAG dynamically switches to graph-based contextualization to recover all structurally related nodes, enabling full evidence coverage.}
    \label{fig:first-case-study}
\end{figure*}

In \Cref{fig:first-case-study}, we present a reference-based question that requires retrieving a multi-page appendix. A standard semantic retriever correctly identifies the node titled ``Appendix A: References'' but fails to locate its continuation on the next page, which lacks semantic similarity to the query. As a result, critical evidence is missed.

LAD-RAG overcomes this by leveraging its symbolic document graph, which encodes layout continuity and section hierarchy. Its dynamic agent follows up on semantic search with graph-based contextualization when additional evidence may extend beyond the initially matched content. Because the document graph already groups nodes into coherent structures during ingestion (via community detection), the retriever can directly exploit these relationships at inference. This allows it to seamlessly retrieve all nodes belonging to the same section or evidence cluster, producing a more coherent and complete set of retrieved evidence.

\medskip
Together, these two cases underscore the unique advantages of LAD-RAG. Rather than relying solely on embedding similarity, it dynamically decides when to use symbolic structure and how to combine it with semantic cues to retrieve comprehensive, contextually grounded evidence from visually complex documents.

\section{Additional Analysis on LAD-RAG's Retrieval Performance}
\label[appsection]{app:additional-analysis-on-retrieval}

\begin{figure}[ht]
    \centering
    \includegraphics[width=\linewidth]{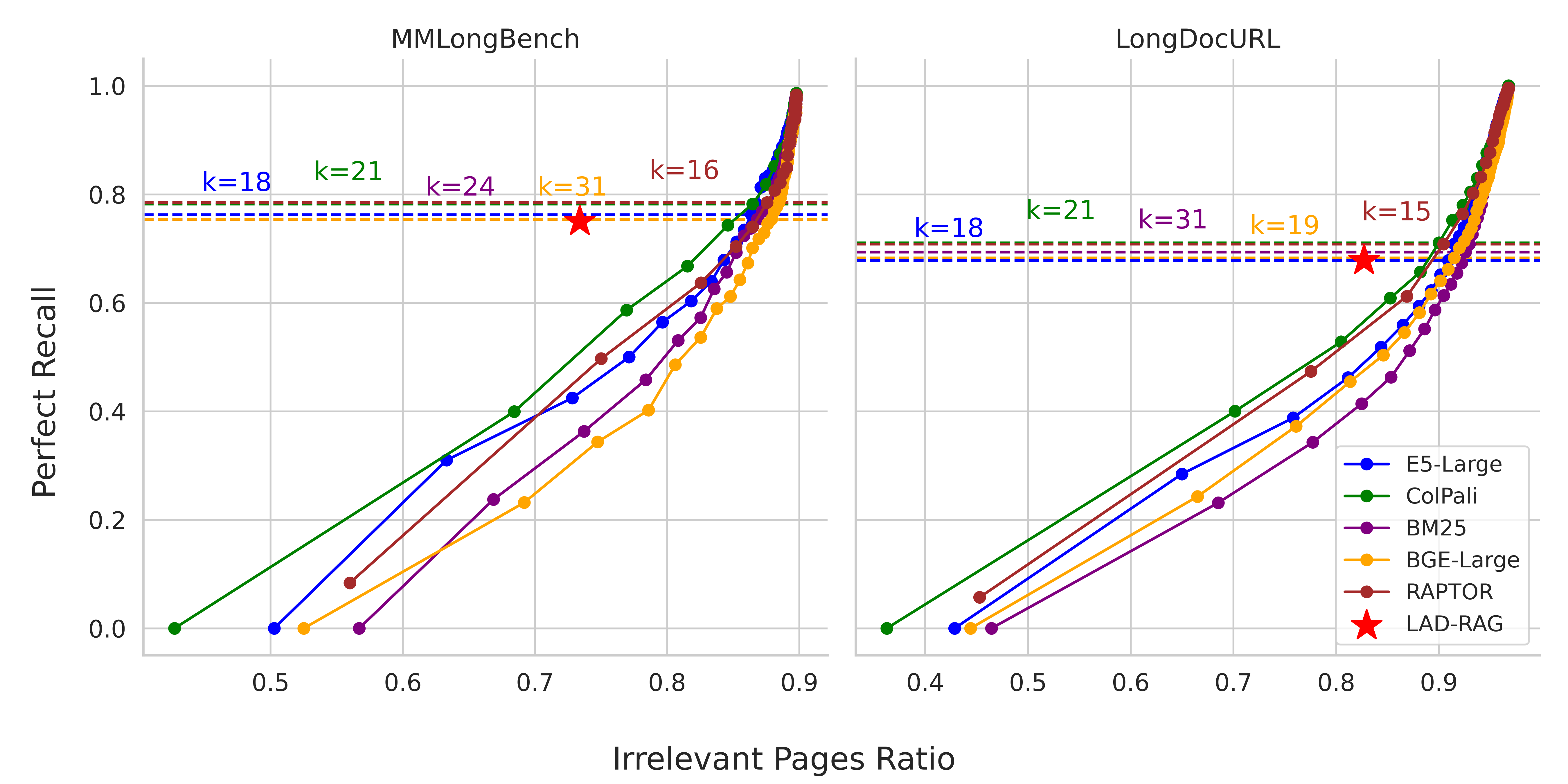}
    \caption{Retrieval performance on multi-page evidence questions. Our framework is compared to baseline retrievers across varying top-$k$ values. Dotted horizontal lines indicate how many pages each baseline must retrieve to match the recall of LAD-RAG without top-$k$ tuning.}
    \label{fig:main-retrieval-multi}
\end{figure}

\Cref{fig:main-retrieval-multi} provides a focused analysis of retrieval performance on multi-page evidence questions, which pose a greater challenge due to the need for cross-page contextual integration. We compare LAD-RAG against conventional top-$k$ retrievers and RAPTOR \citep{sarthi2024raptor}, a hierarchical baseline that groups semantically related elements into aggregate nodes and expands retrievals top-down to their constituents at inference time. While RAPTOR's semantic groupings can help recover related content within locally similar clusters, they do not capture layout continuity or cross-page structural dependencies, making them less effective for multi-page questions where relevant evidence is distributed across structurally related but semantically distant pages. Our method consistently achieves higher recall with fewer irrelevant pages across all baselines, highlighting its ability to prioritize semantically and structurally relevant content across document boundaries. These results underscore the importance of layout-aware, symbolically grounded retrieval strategies for handling complex, distributed information in long documents.

\section{Additional Analysis on LAD-RAG's QA Performance}
\label[appsection]{app:additional-analysis-qa}

\begin{table*}[ht]
\centering
\resizebox{\textwidth}{!}{%
\begin{tabular}{llcccccccc}
\toprule
Model & Retrieval & \multicolumn{5}{c}{Evidence Source} & \multicolumn{3}{c}{Evidence Page} \\
\cmidrule(lr){3-7} \cmidrule(lr){8-10}
 &  & Layout & Text & Table & Figure & Chart & Single & Multi & UNA \\
\midrule
\multirow{5}{*}{InternVL2-8B} & Ground-Truth & 0.385 & 0.427 & 0.315 & 0.380 & 0.339 & 0.506 & 0.250 & 0.052 \\
 & retrieving@5 & 0.256 & 0.298 & 0.266 & 0.199 & 0.295 & 0.372 & 0.164 & \textbf{0.054} \\
 & retrieving@10 & 0.274 & 0.348 & 0.280 & 0.236 & 0.307 & 0.395 & 0.208 & 0.051 \\
 & topk-adjusted & 0.291 & 0.309 & 0.228 & 0.214 & 0.308 & 0.365 & 0.212 & 0.051 \\
 & LAD-RAG & \textbf{0.361} & \textbf{0.408} & \textbf{0.340} & \textbf{0.321} & \textbf{0.363} & \textbf{0.495} & \textbf{0.242} & 0.053 \\
\addlinespace[2pt]
\midrule
\multirow{5}{*}{Pixtral-12B-2409} & Ground-Truth & 0.513 & 0.461 & 0.431 & 0.398 & 0.412 & 0.537 & 0.343 & 0.065 \\
 & retrieving@5 & 0.248 & 0.343 & 0.321 & 0.250 & 0.288 & 0.395 & 0.200 & 0.065 \\
 & retrieving@10 & 0.294 & 0.239 & 0.280 & 0.240 & 0.226 & 0.327 & 0.178 & 0.063 \\
 & topk-adjusted & 0.308 & 0.326 & 0.330 & 0.269 & 0.282 & 0.394 & 0.213 & 0.066 \\
 & LAD-RAG & \textbf{0.336} & \textbf{0.398} & \textbf{0.387} & \textbf{0.316} & \textbf{0.330} & \textbf{0.462} & \textbf{0.261} & \textbf{0.070} \\
\addlinespace[2pt]
\midrule
\multirow{5}{*}{Phi-3.5-Vision} & Ground-Truth & 0.376 & 0.416 & 0.300 & 0.366 & 0.301 & 0.492 & 0.227 & 0.634 \\
 & retrieving@5 & 0.248 & 0.324 & 0.259 & 0.243 & 0.254 & 0.399 & 0.192 & 0.616 \\
 & retrieving@10 & 0.256 & 0.338 & 0.280 & 0.209 & 0.273 & 0.381 & 0.189 & 0.612 \\
 & topk-adjusted & 0.267 & 0.351 & 0.300 & 0.244 & 0.290 & 0.400 & 0.199 & 0.611 \\
 & LAD-RAG & \textbf{0.299} & \textbf{0.364} & \textbf{0.304} & \textbf{0.279} & \textbf{0.324} & \textbf{0.414} & \textbf{0.202} & \textbf{0.632} \\
\addlinespace[2pt]
\midrule
\multirow{5}{*}{GPT-4o-200b-128} & Ground-Truth & 0.597 & 0.628 & 0.638 & 0.536 & 0.616 & 0.693 & 0.565 & 0.206 \\
 & retrieving@5 & 0.378 & 0.469 & 0.500 & 0.395 & 0.435 & 0.607 & 0.303 & \textbf{0.205} \\
 & retrieving@10 & 0.424 & 0.495 & 0.560 & 0.426 & 0.486 & 0.637 & 0.372 & 0.190 \\
 & topk-adjusted & 0.480 & 0.510 & 0.568 & 0.394 & 0.528 & 0.629 & 0.409 & 0.202 \\
 & LAD-RAG & \textbf{0.500} & \textbf{0.576} & \textbf{0.607} & \textbf{0.478} & \textbf{0.534} & \textbf{0.676} & \textbf{0.450} & \textbf{0.205} \\
\bottomrule
\end{tabular}
}
\caption{Breakdown of QA accuracy on \texttt{MMLongBench-Doc}. Best values per model group (excluding \texttt{GT}) are \textbf{boldfaced}.}
\label{tab:mmlongbench_scores}
\end{table*}

\Cref{tab:mmlongbench_scores} provides a fine-grained breakdown of QA accuracy on \texttt{MMLongBench-Doc}, categorized by different evidence sources (e.g., layout, table, figure) and page types (e.g., single-page, multi-page). We observe consistent improvements across all types when using \texttt{LAD-RAG}, with particularly large gains on layout-rich and chart-based questions, highlighting the model's ability to leverage positional and structural cues that are often missed by standard neural retrieval. Improvements are also notable in multi-page contexts, emphasizing the effectiveness of our symbolic document graph in modeling inter-page dependencies that neural retrievers typically fail to capture.

\begin{table*}[t]
\centering
\resizebox{\textwidth}{!}{%
\begin{tabular}{lllrrrrrrr}
\toprule
Model & Retrieval & \multicolumn{4}{c}{Evidence Source} & \multicolumn{3}{c}{Task Type} \\
\cmidrule(lr){3-6}\cmidrule(lr){7-9}
& & Layout & Text & Table & Figure & Und & Rea & Loc \\
\midrule
\multirow{5}{*}{InternVL2-8B} & Ground-Truth & 0.507 & 0.719 & 0.608 & 0.670 & 0.724 & 0.572 & 0.486 \\
 & retrieving@5 & 0.296 & 0.526 & 0.420 & 0.479 & 0.583 & 0.435 & 0.184 \\
 & retrieving@10 & 0.311  & 0.558 & 0.442 & 0.483 & 0.597 & 0.425 & 0.215 \\
 & topk-adjusted & 0.288 & 0.557 & \textbf{0.486} & 0.536 & 0.547 & 0.443 & 0.295 \\
 & LAD-RAG & \textbf{0.319} & \textbf{0.571} & \textbf{0.486} & \textbf{0.556}  & \textbf{0.618} & \textbf{0.625} & \textbf{0.297} \\
\addlinespace[2pt]
\midrule
\multirow{5}{*}{Pixtral-12B-2409} & Ground-Truth & 0.613 & 0.688 & 0.589 & 0.614 & 0.660 & 0.606 & 0.601 \\
 & retrieving@5 & 0.405 & 0.482 & 0.392 & 0.412  & 0.478 & 0.389 & 0.366 \\
 & retrieving@10 & 0.417 & 0.478 & 0.369 & 0.396 & 0.471 & 0.399 & 0.362 \\
 & topk-adjusted & 0.433 & 0.504 & 0.418 & 0.427 & 0.502 & 0.418 & 0.386 \\
 & LAD-RAG & \textbf{0.462} & \textbf{0.553} & \textbf{0.490} & \textbf{0.475} & \textbf{0.559} & \textbf{0.469} & \textbf{0.432} \\
\addlinespace[2pt]
\midrule
\multirow{5}{*}{Phi-3.5-Vision} & Ground-Truth & 0.608 & 0.689 & 0.596 & 0.581 & 0.711 & 0.435 & 0.596 \\
 & retrieving@5 & 0.416 & 0.561 & 0.408 & 0.419 & 0.599 & 0.415 & 0.283 \\
 & retrieving@10 & 0.412 & 0.568 & 0.411 & 0.441 & 0.607 & 0.415 & 0.279 \\
 & topk-adjusted & 0.415 & 0.549 & 0.404 & 0.432 & 0.606 & 0.318 & \textbf{0.298} \\
 & LAD-RAG &  \textbf{0.425} & \textbf{0.577} & \textbf{0.428} & \textbf{0.459} & \textbf{0.619} & \textbf{0.425} & 0.291 \\
\addlinespace[2pt]
\midrule
\multirow{5}{*}{GPT-4o-200b-128} & Ground-Truth & 0.586 & 0.777 & 0.712 & 0.742 & 0.797 & 0.728 & 0.551  \\
 & retrieving@5 & 0.444 & 0.670 & 0.613 & 0.621 & 0.680 & 0.599 & 0.419 \\
 & retrieving@10 & 0.474 & 0.708 & 0.635 & 0.631 & 0.714 & 0.620 & 0.452 \\
 & topk-adjusted & 0.532 & \textbf{0.720} & 0.655 & 0.680 & 0.732 & 0.619 & 0.511 \\
 & LAD-RAG & \textbf{0.536} & 0.718 & \textbf{0.671} & \textbf{0.709} & \textbf{0.735} & \textbf{0.622} & \textbf{0.528} \\
\bottomrule
\end{tabular}
}
\caption{Breakdown of QA accuracy on \texttt{LongDocURL}. Best values per model group (excluding \texttt{GT}) are \textbf{boldfaced}.}
\label{tab:longdocurl_scores}
\end{table*}

A similar pattern is observed in \Cref{tab:longdocurl_scores} for the \texttt{LongDocURL} benchmark. LAD-RAG consistently improves QA accuracy across models and across diverse question types, including understanding, reasoning, and location-based tasks. The gains span questions requiring different forms of document comprehension, from interpreting spatial layout to integrating multi-element evidence, highlighting the versatility of the retrieval strategy. These results suggest that LAD-RAG's benefits are not limited to specific evidence types or query styles, but generalize across a wide range of task demands in visually rich documents.

\Cref{tab:mmlongbench_doc_type_scores} further breaks down QA accuracy on \texttt{MMLongBench-Doc} by document type, including guidebooks, tutorial workshops, brochures, academic papers, research report introductions, financial reports, and administration/industry files. LAD-RAG consistently achieves the best performance across most document types and models, confirming that its benefits generalize across diverse document structures and styles. We observe that academic papers tend to yield lower scores compared to other document types across all retrieval methods. Based on our analysis, this is driven primarily by question difficulty and the reasoning demands placed on the models, rather than by document quality or retrieval failures. LAD-RAG's gains are most pronounced for academic papers, brochures, and administration/industry files, with average improvements of 38\%, 29\%, and 29\% over baseline retrievers respectively. These gains suggest that document types with denser content and more complex cross-page dependencies benefit most from LAD-RAG's symbolic graph-based retrieval.

\begin{table*}[t]
\centering
\resizebox{\textwidth}{!}{%
\begin{tabular}{llccccccc}
\toprule
Model & Retrieval & \multicolumn{7}{c}{Document Type} \\
\cmidrule(lr){3-9}
 & & \makecell{Guide\\book} & \makecell{Tutorial\\Workshop} & Brochure & \makecell{Academic\\Paper} & \makecell{Research Report\\Intro} & \makecell{Financial\\Report} & \makecell{Admin/\\Industry File} \\
\midrule
\multirow{5}{*}{InternVL2-8B} & Ground-Truth & 0.487 & 0.536 & 0.299 & 0.333 & 0.355 & 0.308 & 0.578 \\
 & retrieving@5 & 0.320 & 0.384 & 0.273 & 0.179 & 0.280 & 0.280 & 0.364 \\
 & retrieving@10 & 0.377 & 0.420 & 0.299 & 0.231 & 0.304 & 0.299 & 0.348 \\
 & topk-adjusted & 0.354 & 0.439 & 0.314 & 0.167 & 0.278 & 0.323 & 0.369 \\
 & LAD-RAG & \textbf{0.485} & \textbf{0.588} & \textbf{0.457} & \textbf{0.368} & \textbf{0.446} & \textbf{0.412} & \textbf{0.578} \\
\addlinespace[2pt]
\midrule
\multirow{5}{*}{Pixtral-12B-2409} & Ground-Truth & 0.494 & 0.623 & 0.480 & 0.422 & 0.509 & \textbf{0.419} & 0.580 \\
 & retrieving@5 & 0.397 & 0.460 & 0.365 & 0.314 & 0.409 & 0.350 & 0.444 \\
 & retrieving@10 & 0.387 & 0.464 & 0.380 & 0.191 & 0.385 & 0.325 & 0.284 \\
 & topk-adjusted & 0.408 & 0.482 & 0.391 & 0.286 & 0.408 & 0.364 & 0.395 \\
 & LAD-RAG & \textbf{0.439} & \textbf{0.522} & \textbf{0.430} & \textbf{0.352} & \textbf{0.431} & \textbf{0.419} & \textbf{0.457} \\
\addlinespace[2pt]
\midrule
\multirow{5}{*}{Phi-3.5-Vision} & Ground-Truth & 0.424 & 0.571 & 0.325 & 0.269 & 0.352 & 0.318 & 0.530 \\
 & retrieving@5 & 0.402 & 0.482 & 0.195 & 0.199 & 0.304 & \textbf{0.271} & 0.394 \\
 & retrieving@10 & 0.385 & 0.420 & 0.286 & 0.231 & 0.257 & 0.252 & 0.364 \\
 & topk-adjusted & 0.404 & 0.488 & 0.300 & 0.248 & 0.322 & 0.264 & 0.404 \\
 & LAD-RAG & \textbf{0.424} & \textbf{0.562} & \textbf{0.357} & \textbf{0.269} & \textbf{0.402} & 0.270 & \textbf{0.456} \\
\addlinespace[2pt]
\midrule
\multirow{5}{*}{GPT-4o-200b-128} & Ground-Truth & 0.716 & 0.790 & 0.660 & 0.603 & 0.729 & 0.744 & 0.753 \\
 & retrieving@5 & 0.581 & 0.623 & 0.550 & 0.503 & 0.584 & 0.504 & 0.617 \\
 & retrieving@10 & 0.630 & 0.645 & 0.576 & 0.508 & 0.590 & 0.573 & 0.617 \\
 & topk-adjusted & 0.655 & 0.670 & 0.631 & 0.533 & 0.619 & 0.616 & 0.679 \\
 & LAD-RAG & \textbf{0.658} & \textbf{0.697} & \textbf{0.655} & \textbf{0.550} & \textbf{0.632} & \textbf{0.648} & \textbf{0.744} \\
\bottomrule
\end{tabular}
}
\caption{Breakdown of QA accuracy on \texttt{MMLongBench-Doc} by document type. Best values 
per model group (excluding \texttt{GT}) are \textbf{boldfaced}.}
\label{tab:mmlongbench_doc_type_scores}
\end{table*}

\section{Latency Analysis}
\label[appsection]{app:more-latency}

\begin{figure}[t]
    \centering
    \includegraphics[width=1\linewidth]{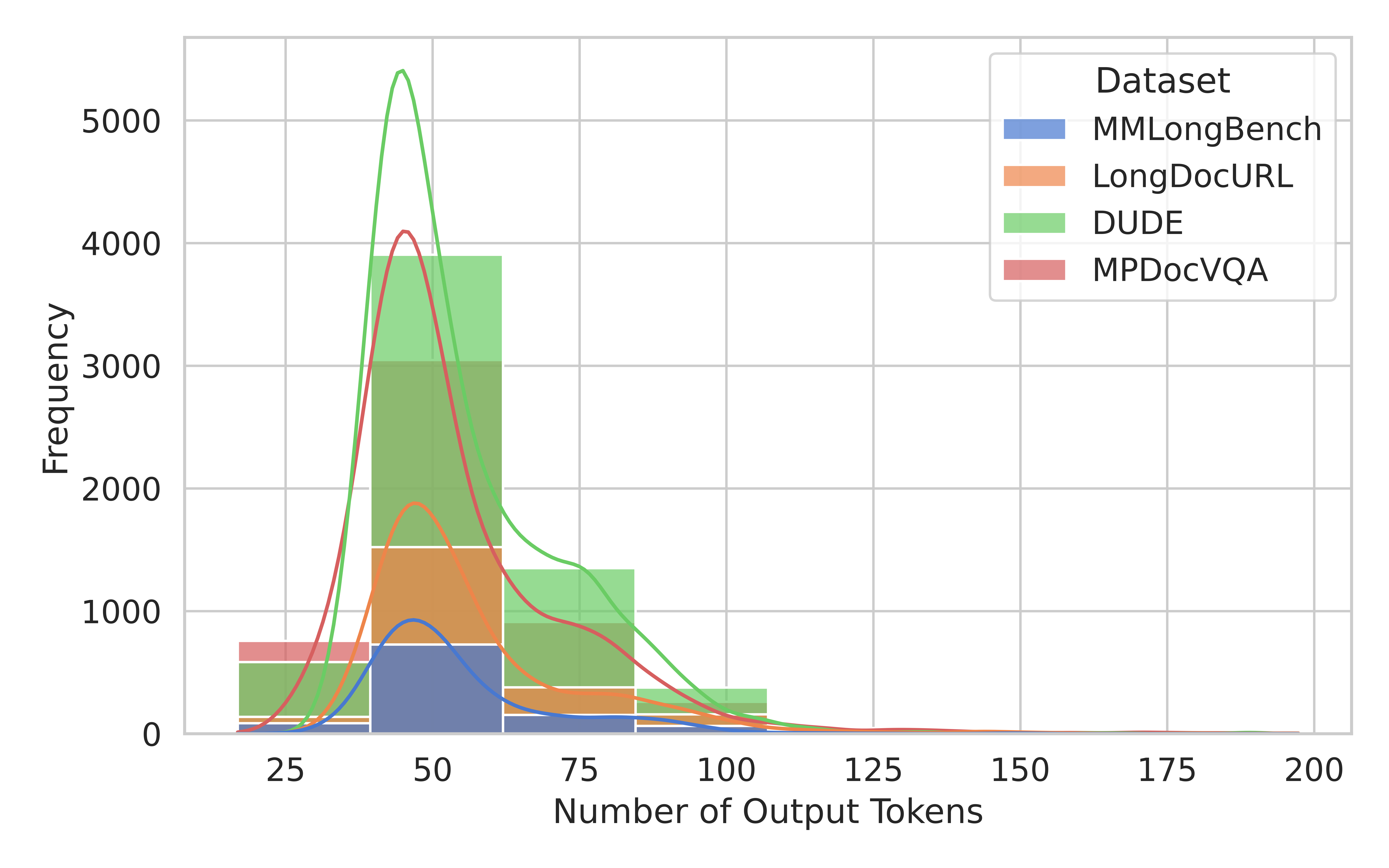}
    \caption{Distribution of tokens generated per query across all LLM calls. The vast majority of queries use fewer than 100 tokens, indicating that the additional reasoning overhead introduced by our dynamic retrieval is lightweight in practice.}
    \label{fig:num_tokens}
\end{figure}

We also measure token generation across all calls. \Cref{fig:num_tokens} shows that more than 97\% of the calls generate fewer than 100 tokens. This distribution highlights that the additional reasoning steps incurred by LAD-RAG are lightweight compared to the overall LLM inference cost. These results, alongside the main results presented in \Cref{subsec:results-latency}, demonstrate that our retrieval strategy introduces only minimal overhead relative to standard RAG pipelines, while delivering significantly higher retrieval completeness and QA accuracy.

\section{Human Evaluation of QA Accuracy}
\label[appsection]{app:human-eval}

To validate our automated QA evaluation pipeline, we sampled 100 examples from LAD-RAG's generated answers on MMLongBench-Doc, balanced between answers labeled as correct and incorrect by our automated judging system. Two human annotators (one computer science graduate student and one computer science undergraduate student) independently judged the correctness of each answer. We then compared their labels against those of the automated judge. Results are shown in \Cref{tab:human-eval}.

\begin{table}[ht]
\centering
\resizebox{\columnwidth}{!}{%
\begin{tabular}{lcc}
\toprule
\textbf{Metric} & \textbf{Accuracy} & \textbf{Cohen's $\kappa$} \\
\midrule
Annotator 1 vs. Judge       & 0.930 & 0.860 \\
Annotator 2 vs. Judge       & 0.950 & 0.900 \\
Annotator 1 vs. Annotator 2 & 0.960 & 0.919 \\
\bottomrule
\end{tabular}
}
\caption{Agreement between human annotators and the automated judge on 100 sampled examples from MMLongBench-Doc. High accuracy and Cohen's $\kappa$ scores indicate strong alignment between human judgments and our automated evaluation pipeline.}
\label{tab:human-eval}
\end{table}

The high agreement scores across both annotator-judge and inter-annotator comparisons confirm that our automated evaluation pipeline reliably reflects human judgment, supporting the validity of the accuracy scores reported throughout the paper.

\section{Implementation Details}
\label[appsection]{app:implementation-details}

We provide all relevant implementation details, including hardware setup, software environment, hyperparameter choices, and prompts used throughout the LAD-RAG framework. The symbolic document graph was implemented using the Python networkx library, which offers rich functionality for representing, analyzing, and traversing graph structures. For model inference, we used the vLLM library (version 0.9.2), which enables fast and efficient LLM serving with minimal overhead. LAD-RAG experiments were conducted on 4 NVIDIA A100 GPUs. The software environment included Python 3.10.12 and PyTorch version \texttt{2.7.0+cu126}.

\subsection{Hyperparameters}

We used temperature $= 0$ for all prompting steps involving the LVLM, including (i) document graph construction (both node and edge generation), (ii) inference-time retrieval via the agent, and (iii) the QA stage. The maximum token limit was set to 8192 for all document graph construction steps, such as page-level element extraction, memory updates, and relation extraction, and 2048 for all QA-related evaluations. For the retriever agent, we set the maximum number of interaction rounds (conversations) to 20. Furthermore, for the baseline retrievers, we evaluate performance across $k$ values up to the point of perfect recall. On average, this occurs at $k{=}94$ for MMLongBench, $k{=}65$ for LongDocURL, $k{=}17$ for DUDE, and $k{=}10$ for MP-DocVQA.

For image rendering, we used the \texttt{PyMuPDF} library to convert each PDF page into an image with a resolution of 300 DPI. During inference, we preserved the original image resolution when GPU memory allowed; in constrained scenarios, we scaled the resolution down to 50\% or, in rare cases, to 20\% of the original size. We manually verified that models could still correctly extract textual and visual information from these downsampled images, ensuring that no critical semantic or layout content was lost.

\subsection{Prompts}
This subsection covers all prompts used with the LVLM across different components of the LAD-RAG framework. This includes prompts for generating document graph nodes and edges during ingestion, as well as the prompts used by the inference-time LLM agent to query the neuro-symbolic index and retrieve contextually relevant evidence to answer questions.

\paragraph{Document graph nodes extraction.}  
We extract layout-aware objects from each document page using a specialized prompt designed to capture rich structural and semantic attributes. The full prompt can be found in \Cref{fig:prompt-docgraph}.

\paragraph{Running memory construction.}  
To build a persistent and structured working memory across multi-page documents, we use a specialized prompt that summarizes, links, and updates previously extracted elements while maintaining coherence and document flow. This prompt helps the system accumulate context and track evolving sections or entities across pages. The complete prompt is shown in \Cref{fig:prompt-memory}.

\paragraph{Document graph generation.}
We provide the full prompt used to construct the document graph during ingestion, with a focus on modeling cross-page structure and dependencies.
This prompt guides the model to maintain a structured working memory across pages, update hierarchical and semantic state, and extract relationships that connect document elements (e.g., paragraphs, tables, figures) across page boundaries.
It instructs how to represent elements as graph nodes and link them using edges encoding layout-aware, semantic, and interpretive relationships spanning multiple pages.
The full text of the prompt is shown in \Cref{fig:prompt-graph-construction}.

\paragraph{Retriever agent.}
We show the full prompt used to guide the retriever agent during inference. This agent is responsible for orchestrating multiple retrieval rounds based on partial observations and evolving context. The prompt describes how the agent should form hypotheses, refine queries, decide when to stop, and integrate symbolic and semantic cues to gather the most relevant evidence. See \Cref{fig:prompt-retriever-agent} for the complete prompt.

\newpage
\begin{figure*}[ht]
\centering
\begin{fullpromptbox}[Prompt used for document graph node extraction]
\begin{Verbatim}
You are a document/image analysis assistant. The user is uploading a scanned or rendered page from a PDF file or it can even be an image. Your task is to extract a structured JSON representation of the page/image, breaking it down into individual, self-contained objects such as paragraphs, titles, section headers, tables, and figures.
For each detected object, extract the following fields:
- `type`: One of [title, section_header, paragraph, figure, table, footnote, metadata, etc.]
- `content`: The full raw content of the object. This must include **all information visible** in the element, 
  whether it's a paragraph, table, or chart. For tables or figures, include captions, all data values, axis labels, 
  legend descriptions, etc.
- `title_or_heading`: A short heading that describes what this object is about.
- `position_on_page`: Where this object appears (e.g., 'top-left', 'center-bottom', 'right margin').
- `layout_relation`: How this object is situated relative to other elements (e.g., 'below the main title', 
  'next to bar chart', 'in footer section').
- `summary`: A **complete and self-contained** summary of the object. Summarize all meaningful information in the object. 
  For example, for tables or charts, describe the key patterns, values, and implications clearly. The summary should be 
  useful even when this object is retrieved alone in a retrieval-augmented generation (RAG) system.
- `visual_attributes`: Font size, emphasis (bold/italic), color, spacing, etc., if observable.
- `page_metadata`: Metadata such as `document_title`, `page_number`, `watermark`, `source_url`, if available.
- `object_id`: A unique ID (e.g., '{}-obj_003') that can be used to reference this object in a downstream system. 
  All the IDs should start with the prefix '{}'. DO NOT ADJUST THIS PREFIX WITH THE PAGE NUMBER ON THE PAGE AND JUST USE THE PREFIX GIVEN TO YOU.
- (optional) `content_type`: A semantic tag like 'contact_info', 'citation', 'header', 'statistical_summary', etc., if applicable.
- (optional) `document_context`: A brief reference to the broader document or section context (e.g., 
  'From Pew Research Center report on U.S. immigration opinion, May 2018').
Important:
- Each object must be maximally self-contained and descriptive.
- The `content` field should include all information that appears in the visual object.
- The `summary` should explain the meaning, relevance, and insight of the object, not just repeat its text.
- Your output must be a **JSON list**, where each item is an object containing the fields listed above. 
- If any field is unknown or unavailable, return null.
- The result must be optimized for downstream use in a RAG system.
\end{Verbatim}
\end{fullpromptbox}
\caption{Prompt used for document graph node extraction.}
\label{fig:prompt-docgraph}
\end{figure*}

\newpage
\begin{figure*}[ht]
\centering
\begin{fullpromptbox}[Prompt used for running memory construction]
\begin{Verbatim}
You are analyzing the structure of a multi-page document. Your task is to decide how to update the document’s `section_queue` based on a list of new section-like objects detected on the current page.

The `section_queue` represents the current hierarchical structure of the document. It is a list that moves from the top-level section to deeper subsections. Each element is either:
- a single dictionary with `text` and `object_id` representing a section/subsection.

Each dictionary has the form:
- `"text"`: the section or subsection title
- `"object_id"`: the ID of the corresponding object

Now, for the current page, you are given `candidate_sections`, a list of newly extracted section-like objects. Your job is to determine:
- If these candidates are **new subsections** under the current section (append them).
- If the new candidates are semantically on the same levels as the leaf already in the current hierarchy, truncate the tree until you get to a broader topic and then insert them. Sometimes you might need to remove all the topic and start from scratch, which is OK. 
- Or if they are just **continuing** the existing structure (no change).

You should use **semantic understanding** of the section titles to determine structural relationships — for example, whether “Threats to Democracy” fits under “Challenges in Governance” or signals a new section.

---

**CURRENT SECTION_QUEUE:**
{json.dumps(working_memory.get('section_queue', working_memory), indent=2)}

**CANDIDATE SECTIONS FROM PAGE:**
{json.dumps(candidate_section_objects, indent=2)}

Return only a JSON object with the updated section queue:
```json
{{
  "section_queue": [
    {{ "text": ..., "object_id": ... }},
    ...
    [{{ "text": ..., "object_id": ... }}, {{ "text": ..., "object_id": ... }}],
  ]
}}
Important rules:
- Only return the final updated section_queue JSON — no explanations or extra texts.
- It SHOULD NEVER happen that objects lower in the hierarchy and closer to the leaf would come from pages prior to the parent objects (this can be inferred from the object ids: ...page_{{ page number}}...). If you're updating the section_queue like that, something is wrong; Do it again. objects closer to the leafs are both structurally (page-wise) and also semantically following parent objects and NOT the other way around.
\end{Verbatim}
\end{fullpromptbox}
\caption{Prompt used for constructing and updating the running memory across document pages.}
\label{fig:prompt-memory}
\end{figure*}

\newpage
\begin{figure*}[ht]
\centering
\begin{fullpromptbox}[Prompt used for document graph construction]
\begin{Verbatim}
You are continuing the analysis of a document with multiple pages. Below is the content extracted from the current page, including structured objects and intra-page relationships, as well as the actual document page which is given to you.

Use this information along with your memory of prior pages and the structural information obtained so far to:

- **Update the working memory**. The working memory should maintain both semantic and structural state across pages, including:
  • **section_queue**: a hierarchical list of sections and subsections, possibly including lists of sibling sections at the same level. Do **not** change this unless explicitly asked; it has already been updated.
  • **active_entities**: a list of important entities that persist across pages, each with `text` and associated `object_id(s)`
  • **semantic_topics**: any core topics being discussed that continue across pages
  • **unresolved_objects**: any objects (e.g., tables, figures) that are referred to but not fully explained yet (with `object_id`)

- **Identify cross-page relationships** between this page and earlier pages. This includes:

  1. **Hierarchical references**:
     - All extracted objects from the current page (e.g., paragraphs, figures, tables) should be **explicitly connected** to the most relevant section or subsection from the current `section_queue`, prioritizing the lower leaf levels. 
     - If there are multiple valid target sections or subsections (e.g., leaf or its parent in the hierarchy), use both.
     - Represent these using:
       {{
         "from_object": "object_id_of_child",
         "to_object": "object_id_of_section",
         "type": "is_part_of_section"
       }}

  2. **Interpretive links**:
     - For example, if a paragraph on this page explains a table from a previous page, return:
       {{
         "from_object": "paragraph_obj_id",
         "to_object": "table_obj_id",
         "type": "explains"
       }}

  3. **Object-level semantic links**:
     - Link the current page’s objects to earlier objects referenced in working memory using semantic types such as:
       `"continues"`, `"references"`, `"summarizes"`, `"updates"`, ...
     - These links should be grounded in the objects tracked in the working memory (e.g., those listed in `active_entities`, `semantic_topics`, and `unresolved_objects`).

CURRENT PAGE OBJECTS:
{extracted_objects_text}

CURRENT PAGE RELATIONSHIPS:
{extracted_relations_text}

CURRENT WORKING MEMORY:
{json.dumps(working_memory, indent=2)}


Return a JSON object with the following fields:
- "updated_memory": updated version of the working memory
- "cross_page_relationships": a list of new relationships across pages in the format:
  {{
    "from_object": "object_id_of_child_or_reference",
    "to_object": "object_id_of_parent_or_target",
    "type": "relationship_type"
  }}
\end{Verbatim}
\end{fullpromptbox}
\caption{Prompt used for document graph construction during ingestion.}
\label{fig:prompt-graph-construction}
\end{figure*}

\newpage
\begin{figure*}[ht]
\centering
\begin{fullpromptbox}[Prompt used for retriever agent inference]
\begin{Verbatim}
<task_description>
You are an agent designed to retrieve evidence from a visually-rich document using a structured document graph.
Your goal is to dynamically determine and retrieve the best set of evidence nodes based on the question.
Do NOT try to filter too much based on small details and as much as possible try to prioritize recall over precision. for instance, if a question is asking about all the charts or figures with certain characteristic, get all the figures and return them as when you're doing extra filtering you might lose important evidences.
You can request code to operate over either:
    - A semantic similarity index (for conceptually close matches),
    - A graph-based document structure (for layout-aware or structural reasoning),
    - Or a combination of both.
The retrieval process is iterative. First, create a high-level plan of what operations are required (numbered), and then for each step, generate a Python code snippet. I will execute each code snippet and return the results, until you determine that sufficient evidence has been retrieved.
</task_description>
<data_description>
Each evidence comes from a node in an undirected NetworkX document graph with the following structure; Not all keys are guaranteed to exist—code should handle missing fields gracefully.
<node>
    <type>One of: aggregated_section, paragraph, section_header, title, figure, metadata, table, footnote, footer, list, note, quote, citation, link, contact_info, code, warning, map, page_number, code_block</type>
    <content>Full text of the node</content>
    <summary>Natural language summary of content</summary>
    <object_id>Unique node identifier which can have the format [main folder]/[document]/page_[page number]-obj_[object number]; 
    </object_id>
</node>
</data_description>
<available_operations>
You can use the following operations in your plan:
1. Provide one string query to retrieve top-k semantically similar nodes.
2. Use structural filters to directly select nodes based on their attributes.
3. Given a node's object_id, return related content based on graph structure. This operation provides contextualization for nodes, so try to use it as much possible.
</available_operations>
<task_flow>
1. Read the <question> and understand what kind of evidence it requires (e.g., something more semantic or layout-related, e.g., figure? table? all sections on a page?).
2. Propose a <plan> with numbered steps and a short description of each.
3. For each step in the plan, generate valid Python code.
4. Wait for me to execute and return the result.
5. Iterate as needed until all relevant evidence is retrieved and do not filter too much; again recall is more important than precision.
7. Finish with a list of selected node IDs in <code></code> and <step_keyword>DONE</step_keyword>. 
</task_flow>
<output_format>
Always start with:
    <plan>...</plan>
Then, for each step, which is also a turn in our conversation:
    <step>
        <description>Short goal of this step</description>
        <step_keyword>from the set {{semantic_search,graph_filter,graph_contextualize,DONE}}</step_keyword>
        <code>
            # Python code here
        </code>
    </step>
**for semantic_search, you have access to the function do_semantic_search(query, {}) that gets a query in string and the pdf_name that you do NOT need to change, and retrieves the most relevant nodes matching wrt to their summary and content for that string. you can do this multiple times if you think you need to retrieve the semantically similar items for multiple queries that are part of the main query. Be sure to only give me the function call; I will put the result of the function call in a variable and give it back to you myself.
**for graph_filter, you have access to a doc_graph that is basically the networkX object we talked about. You should give me the script that works with this doc_graph and its content is a list of nodes after filtering; Be sure to give me a expression not a statement; for instance [(node_id, node) for node_id, node in doc_graph.nodes(data=True) if ...]. 
**for graph_contextualize, you have access to the function get_community_for_node(node_id, doc_graph) that basically gets all the nodes that are in the community of that specific node. Be sure to only give me the function call; I will put the result of the function call in a variable and give it back to you myself.
</output_format>
<instructions>
You are now ready to begin. Think step by step and act like a smart retrieval agent.
Respond by first outputting a <plan> based on the following question: "{}"
</instructions>
\end{Verbatim}
\end{fullpromptbox}
\caption{Prompt used to guide the retriever agent during dynamic iterative evidence retrieval.}
\label{fig:prompt-retriever-agent}
\end{figure*}



\end{document}